\documentclass{article}

\usepackage[final]{neurips_data_2022}
\usepackage[utf8]{inputenc} 
\usepackage[T1]{fontenc}    
\usepackage{hyperref}       
\usepackage{url}            
\usepackage{booktabs}       
\usepackage{amsfonts}       
\usepackage{nicefrac}       
\usepackage{microtype}      
\usepackage{xcolor}         

\usepackage{color}
\usepackage{graphicx}
\usepackage{multicol}
\usepackage{multirow}
\usepackage{array}
\usepackage{xspace}
\usepackage{enumitem}
\setlist[itemize]{leftmargin=*}
\setlist[enumerate]{leftmargin=*}
\usepackage{amsmath}

\newcommand{\nop}[1]{}

\newcommand{\rebuttal}[1] {{{#1}}}

\newcommand{\methodFont}{\textsl}

\newcommand{\bt}{\methodFont{BT}\xspace}

\newcommand{\ie}{\textit{i.e.}\xspace}
\newcommand{\eg}{\textit{e.g.}\xspace}
\newcommand{\etc}{\textit{etc}\xspace}

\newcommand{\hok}{\textit{Honor of Kings}\xspace}
\newcommand{\hokenv}{\textit{Honor of Kings Arena}\xspace}

\usepackage{newfloat}
\usepackage{listings}

\usepackage{amsfonts,amsthm,amsmath}
\usepackage{braket}
\usepackage{balance}
\usepackage[ruled,linesnumbered]{algorithm2e} 
\usepackage{booktabs}
\usepackage[font=small,labelfont=bf]{caption}
\usepackage{marvosym}
\usepackage{ifsym}

\definecolor{codegreen}{rgb}{0,0.6,0}
\definecolor{codegray}{rgb}{0.5,0.5,0.5}
\definecolor{codepurple}{rgb}{0.58,0,0.82}
\definecolor{backcolour}{rgb}{0.95,0.95,0.92}
\lstdefinestyle{mystyle}{
    backgroundcolor=\color{backcolour},   
    commentstyle=\color{codegreen},
    keywordstyle=\color{magenta},
    numberstyle=\tiny\color{codegray},
    stringstyle=\color{codepurple},
    basicstyle=\ttfamily\footnotesize,
    breakatwhitespace=false,         
    breaklines=true,                 
    captionpos=b,                    
    keepspaces=true,                 
    numbers=left,                    
    numbersep=5pt,                  
    showspaces=false,                
    showstringspaces=false,
    showtabs=false,                  
    tabsize=2
}

\makeatletter
\def\thanks#1{\protected@xdef\@thanks{\@thanks
        \protect\footnotetext{#1}}}
\makeatother
\def\BibTeX{{\rm B\kern-.05em{\sc i\kern-.025em b}\kern-.08emT\kern-.1667em\lower.7ex\hbox{E}\kern-.125emX}}

\title{Honor of Kings Arena:
an Environment for Generalization in Competitive Reinforcement Learning}

\author{%
  Hua Wei$^{* \dag  \flat }$, Jingxiao Chen$^{*\ddag \flat}$, Xiyang Ji$^{ *\S}$\thanks{\textsuperscript{*} Authors contributed equally; \textsuperscript{$\flat$} work done at Tencent}, Hongyang Qin$^{\S}$, Minwen Deng$^{\S}$, Siqin Li$^{\S}$,  \\
  \textbf{Liang Wang$^{\S}$, Weinan Zhang$^\ddag$, Yong Yu$^\ddag$, Lin Liu$^\natural$, Lanxiao Huang$^\natural$, }\\
  \textbf{Deheng Ye$^{\S}$\textsuperscript{\Letter \thanks{\textsuperscript{\Letter} Corresponding author}}, Qiang Fu$^{\S}$, Wei Yang$^{\S}$} \\
  $^{\S}$Tencent AI Lab, $^{\natural}$Tencent Timi Studio, \\ $^{\dag}$New Jersey Institute of Technology, $^{\ddag}$Shanghai Jiao Tong University\\
hua.wei@njit.edu, timemachine@sjtu.edu.cn, wnzhang@sjtu.edu.cn, yyu@apex.sjtu.edu.cn, \\
\{xiyangji, hongyangqin, danierdeng, gracesqli, enginewang, lincliu, jackiehuang, \\
dericye, leonfu, willyang\}@tencent.com \\
}

\begin{document}

\maketitle

\begin{abstract}
This paper introduces \hokenv, a reinforcement learning (RL) environment based on \hok, one of the world's most popular games at present. Compared to other environments studied in most previous work, ours presents new generalization challenges for competitive reinforcement learning. It is a multi-agent problem with one agent competing against its opponent; 
and it requires the generalization ability as it has diverse targets to control and diverse opponents to compete with. We describe the observation, action, and reward specifications for the \hok domain and provide an open-source Python-based interface for communicating with the game engine. We provide twenty target heroes with a variety of tasks in \hokenv and present initial baseline results for RL-based methods with feasible computing resources. Finally, we showcase the generalization challenges imposed by \hokenv and possible remedies to the challenges. All of the software, including the environment-class, are publicly available at: \url{https://github.com/tencent-ailab/hok_env}.  
The documentation is available at: \url{https://aiarena.tencent.com/hok/doc/}. 
\end{abstract}

\section{Introduction}

Games have been used as testbeds to measure AI capabilities in the past few decades, from backgammon~\cite{tesauro1994td} to chess~\cite{silver2017mastering} and Atari games~\cite{mnih2013playing}. In 2016, AlphaGo defeated the world champion through deep reinforcement learning and Monte Carlo tree search~\cite{silver2017mastering}. In recent years, reinforcement learning models have brought huge advancements in robot control~\cite{haarnoja2018soft}, autonomous driving~\cite{pan2017virtual}, and video games like StarCraft~\cite{alphastarmastering}, Dota~\cite{berner2019dota}, Minecraft~\cite{guss2019minerl} and \hok~\cite{wu2019hierarchical,ye2020towards,ye2020mastering}.


Related to previous AI milestones, the research focus of game AI has shifted from board games to more complex games, such as imperfect information poker games~\cite{bowling2015heads} and real-time strategic games~\cite{ye2020towards}. As a sub-genre of real-time strategic games, Multi-player Online Battle Arena (MOBA) games have attracted much attention recently~\cite{berner2019dota,ye2020mastering}. The unique playing mechanics of MOBA involve role/hero play and multi-player. Especially since MOBA games have different roles/heroes and each role has different actions, a good AI model needs to perform stably well in controlling the actions of different heroes against different opponent heroes. This makes MOBA 1v1 games, which focus on hero control \cite{ye2020mastering}, a perfect testbed to test the generality of models under different tasks.

Existing benchmark environments on RL generality \rebuttal{are} mainly focusing on relatively narrow tasks for a single agent. \rebuttal{For example, MetaWorld~\cite{yu2020meta} and RLBench~\cite{james2020rlbench}} present benchmarks of simulated manipulation tasks in a shared, table-top environment with a simulated arm, whose goal is to train the arm controller to complete tasks like opening the door, fetching balls, \etc. As the agent's action space remains the same as an arm, it is hard to tell the generality of the learned RL on more diverse tasks like simulated legs.

\begin{figure*}[t!]
\small
\centering
  \begin{tabular}{ccc}
  \includegraphics[height=.21\linewidth]{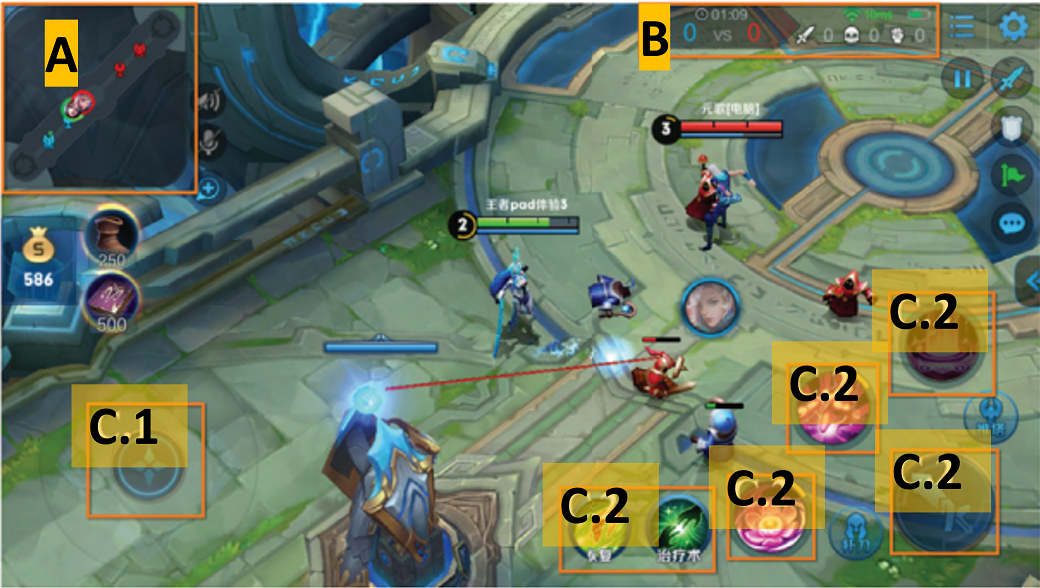} &
  \includegraphics[height=.21\linewidth]{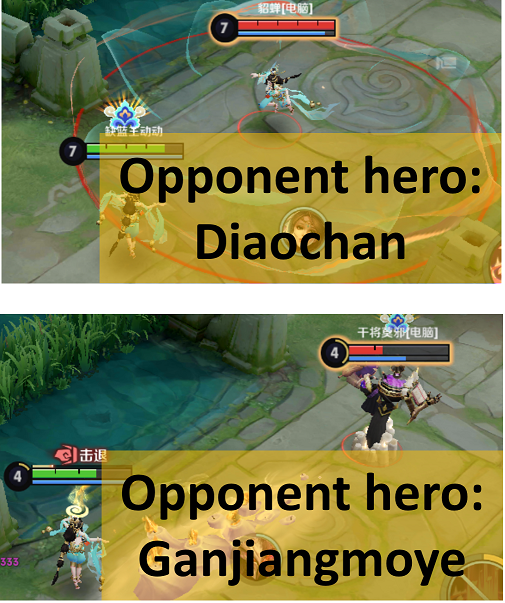} &
  \includegraphics[height=.21\linewidth]{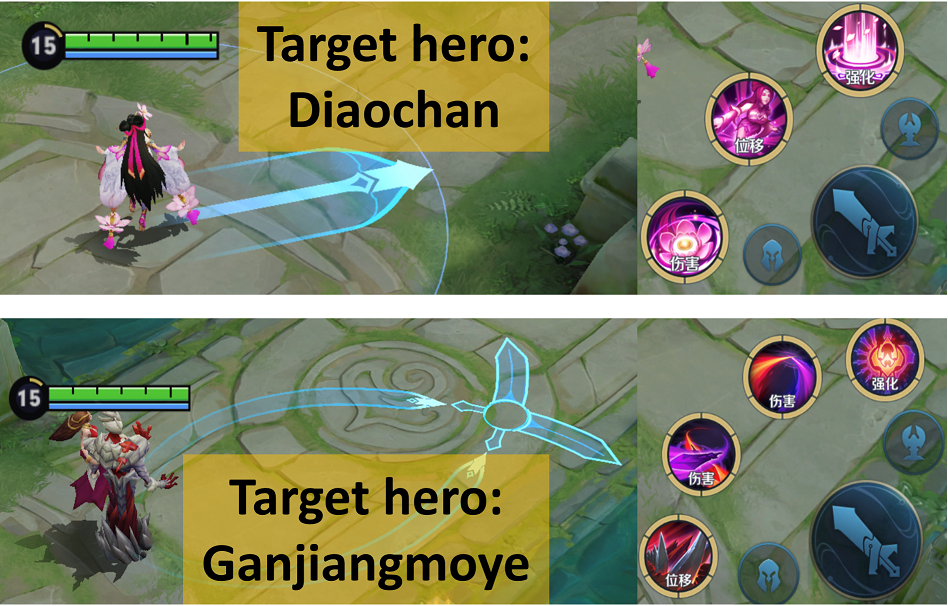}\\
  (a) User interface & (b) Change of opponents & (c) Change of targets  \\
  \end{tabular}
    \vspace{-3mm}
 \caption{Game user interface (UI) and the change of opponents and targets in one match of \hok. (a) In the main screen, there are four sub-parts: a mini-map on the top-left, a dashboard that records the number of KDAs (kill/death/assist) on the top-right, a movement controller on the bottom-left, and skill controller buttons on the bottom-right. (b) The environment changes with different opponent heroes. (c) The action space changes with different target heroes.}
 \vspace{-5mm}
    \label{fig:gen}
\end{figure*}

In this paper, we provide \hokenv, a MOBA 1v1 game environment, authorized by the original game \hok  \footnote{\url{https://en.wikipedia.org/wiki/Honor_of_Kings}}.  The game of \hok was reported to be one of the world's most popular and highest-grossing games of all time, as well as the most downloaded App worldwide. As of November 2020, the game was reported to have over 100 million daily active players~\cite{onemillion}.  
There are two camps in MOBA 1v1, each with one agent, and each agent controls a hero character. As shown in Figure~\ref{fig:gen}(a), an \hok player uses the bottom-left steer button to control the movements of a hero and uses the bottom-right set of buttons to control the hero's skills. 
To win a game, agents must take actions with planning, attacking, defending and skill combos, with consideration on the opponents in the partially observable environment. 

Specifically, the \hokenv imposes the following challenges regarding generalization:
~\noindent\\~$\bullet$~\textbf{Generalization across opponents}. When controlling one target hero, its opponent hero varies across different matches. There are over 20 possible opponent heroes in \hokenv(in the original game there are over 100 heroes), each having different influences in the game environment. If we keep the same target hero and vary the opponent hero as in Figure~\ref{fig:gen}(b), \hokenv could be treated as a similar environment as MetaWorld~\cite{yu2020meta}, which both provides a variety of tasks for the same agent with the same action space. 
~\noindent\\~$\bullet$~\textbf{Generalization across targets}. The generality challenge of RL arises to a different dimension when it comes to the competitive setting. In a match of MOBA game like \hok and DOTA, players also need to master different hero targets. Playing a different MOBA hero is like playing a different game since different heroes have various attacking and healing skills, and the action control can completely change from hero to hero, as shown in Figure~\ref{fig:gen}(c). With over 20 heroes to control for \hokenv, it calls for robust and generalized modeling in RL.

\textbf{Contributions:} As we will show in this paper, the above-mentioned challenges are not well solved by existing RL methods under \hokenv. In summary, our contributions are as follows:
~\noindent\\~$\bullet$~ We provide the \hokenv, a highly-optimized game engine that simulates the popular MOBA game, \hok. It supports 20 heroes in the competitive mode.
~\noindent\\~$\bullet$~ We introduce simple and standardized APIs to make RL in \hok straightforward: the complex observations and actions are defined in terms of low-resolution grids of features; configurable rewards are provided combining factors like the score from the game engine.
~\noindent\\~$\bullet$~ We evaluate RL algorithms under \hokenv, providing an extensive set of benchmarking results for future comparison.
~\noindent\\~$\bullet$~ The generality challenges in competitive RL settings are proposed with preliminary experiments showing that existing RL methods cannot cope well under \hokenv.


\textbf{What \hokenv isn't:} \hokenv currently only supports the 1v1 mode of \hok, where there is only one hero in each camp. Though we acknowledge that some game modes of \hok (\ie, 3v3 or 5v5 mode where there are multiple heroes in each camp) are more popular than 1v1 mode, they might complicates the generalization challenge in competitive RL with the cooperation ability between heroes (which is out of the scope of this paper). The cooperation ability also makes it hard to evaluate the generalization ability of different models. Thus, \hokenv leaves these 3v3 and 5v5 mode of \hok out of the current implementation.
\section{Motivations and Related Work}


The key motivation behind \hokenv is to address a variety of needs that existing environments lack:
\vspace{-3mm}
\noindent \paragraph{Diversity in controlled targets} 
A unique feature of \hokenv is that it has 20 heroes for the agents to control, where each hero has its unique skills. In most existing open environments like Google Research Football~\cite{kurach2020google}, StarCraft2 AI Arena~\cite{samvelyan2019starcraft}, Blood Bowl~\cite{justesen2019blood}, MetaWorld~\cite{yu2020meta} \rebuttal{and RLBench~\cite{james2020rlbench}}, the meaning of actions remains the same when the agent controls different target units. The change of action control between a great number of heroes in \hokenv provides appealing scenarios for testing the generalization ability of the agents. \rebuttal{RoboSuite~\cite{zhu2020robosuite} provides seven different robotic arms across five single-agent tasks and three cooperative tasks between two arms. HoK environment differs from Robosuite by providing a large number of competitive settings.}

\vspace{-3mm}
\paragraph{Free and open accessibility}  DotA2 shares a similar game setting with \hok (both are representative MOBA games with large state/action spaces), which also has multi-agent competition and coordination, unknown environment model, partially observable environment and diverse target heroes to control, but the environment in~\cite{berner2019dota} used by OpenAI is not open with only an overview posted. 
Other MOBA environments like Derk's Gym~\cite{gym_derk} lack free accessibility because of the requirement of commercial licenses. \rebuttal{ XLand~\cite{team2021open} also focuses on the generalization capability of agents and supports multi-agent scenarios, but it is not open-source.}
\vspace{-3mm}
\paragraph{Existing Interest}
This environment has been used as a testbed for RL in research competitions~\footnote{\url{https://aiarena.tencent.com/aiarena/en}} and many researchers have conducted experiments under the environment of Honor of Kings~\cite{chen2021heroes,cheng2019makes, jiang2018feedback,wang2019unsupervised,wei2021boosting,wu2019hierarchical,ye2020towards,ye2020mastering,ye2020supervised}.
Though some of them verified the feasibility of reinforcement learning in tackling the game~\cite{jiang2018feedback,wu2019hierarchical,ye2020towards,ye2020mastering}, they are more focused on methodological novelty in planning, tree-searching, \etc.
Unlike these papers, this paper focuses on making the environment open-accessible and providing benchmarking results, which could serve as a reference and foundation for future research. 
Moreover, this paper showed the weaknesses of former methods in lacking of model generalization across multiple heroes.

\section{\hokenv Environment}
\label{sec:env}

\hokenv is open-sourced under Apache License V2.0 and \rebuttal{accessible to all individuals for any non-commercial activity.} 
The encrypted game engine and game replay tools \rebuttal{follows Tencent's Hornor of Kings AI And Machine Learning License~\footnote{\url{https://github.com/tencent-ailab/hok_env/blob/master/GAMECORE.LICENSE}} and} can be downloaded from: \url{https://aiarena.tencent.com/hok/download}. 
The code for agent training and evaluation is built with official authorization from \hok and is available at: ~\url{https://github.com/tencent-ailab/hok_env}. 
\rebuttal{Any non-commercial users are free to download our game engine and tools after registration}. 


   

   

\subsection{Tasks}
We use the term "task" to refer to specific configurations of an environment (e.g., game setting, specific heroes, number of agents, etc.).
The general task for agents in \hokenv is as follows: When the match starts, each player controls the hero, sets out from the base, gains gold and experience by killing or destroying other game units (e.g., enemy heroes, creeps, turrets). The goal is to destroy the opponent's turrets and base crystal while protecting its own turrets and base crystal. A detailed description of the game units and heroes can be found in Appendix~\ref{append:basic}.

\begin{figure*}[t!]
\centering
  \includegraphics[width=.95\linewidth]{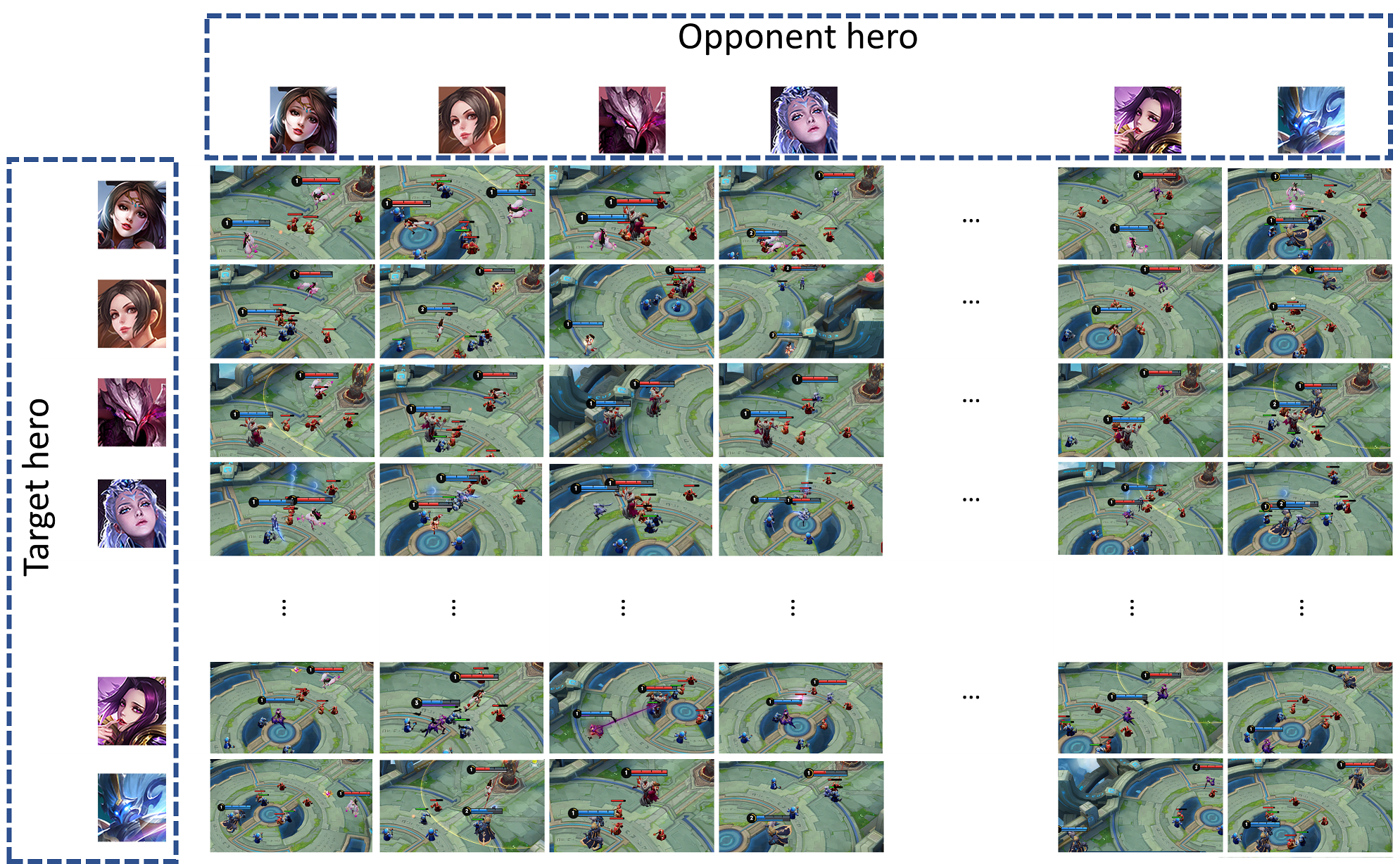}
    \vspace{-1mm}
 \caption{The tasks in \hokenv. Each row represents the same target hero with different opponent heroes. Each column represents different target heroes with the same opponent hero. There are 20 heroes in \hokenv, making $20\times 20=400$ tasks in total.}
    \label{fig:tasks}
    \vspace{-4mm}
\end{figure*}

Though the general goal is the same across different matches, every match would differentiate from each other. Before the match starts, each player needs to choose one hero to control, where each hero has its unique skills, which would have different influences on the environment. Any changes in the chosen hero would make the task different. As shown in Figure~\ref{fig:tasks}, in the current \hokenv, 20 heroes could be chosen, which makes up $400$ tasks in total. 

\subsection{Agents}\label{sec:agents}
\hokenv provides recognizable and configurable observation spaces, action spaces, and reward functions. In this section, we provide a general description of these functions, whose details can be found in Appendix. 
\vspace{-3mm}
\paragraph{Observation Space} The observations often carry spatial and status cues and suggest meaningful actions to perform in a given state. In \hokenv, the observation space is designed to be the same across all heroes, creating the opportunity to generalize across tasks. Specifically, the observation space of \hokenv consists of five main components, whose dimensions depend on the number of heroes in the game (for the full description, please see Appendix~\ref{append:herostate}):
\lstinline|HeroStatePublic|,  which describes the hero's status; 
\lstinline|HeroStatePrivate|, which includes the specific \rebuttal{skill} information for all the heroes in the game; \lstinline|VecCreeps| describing the status of soldiers in the troops; \lstinline|VecTurrets| describing the status of turrets and crystals; \lstinline|VecCampsWholeInfo|, which indicates the period of the match. 
\vspace{-3mm}
\paragraph{Action Space}
The native action space of the environment consists of a triplet form, \rebuttal{which covers all the possible actions of the hero hierarchically:  1) which action button to take; 2) who to target,\eg, a turret, an enemy hero, or a soldier in the troop; 3) how to act, \eg, the discretized direction to move and release skills.} Note that different heroes have different prohibited skill offsets since they have different skills.

\vspace{-3mm}
\paragraph{Reward Information}
\hok has both sparse and dense reward configurations in five categories: farming related,  kill-death-assist (KDA) related, damage related, pushing related, and win-lose related (for the full description, please see Appendix~\ref{append:rewarddesign}). 

\vspace{-3mm}
\paragraph{Episode Dynamics}
An episode of \hokenv task terminates when the crystal of one camp is pushed down. In practice, there is a time limit in training, though an actual round of \hok game has no time limit. The timer is set at the beginning of the episode. The actions in \hokenv are executed every 133ms by default to match with the response time of top-amateur players, while the action interval is configurable. The constraints of the game are expressed in system state transitions of the game. For example, the HP of opponent's crystal will not decrease if the turret is not destroyed. 


\section{APIs and Implementation}
\label{sec:api}

\paragraph{The RL Environment Class}
The class \lstinline{hok1v1}, which can be found within the \lstinline{HoK} module, defines two important functions \lstinline{reset()} and \lstinline{step()}, for starting a new episode, and for advancing time given an action, respectively. 
They both return a quadruple:
\begin{itemize}[leftmargin=20pt]
 \item \lstinline{obs}: a list of NumPy arrays with the description of the agent's observation on the environment.
 \item \lstinline{reward}: a list of scalars with the description of the immediate reward from the environment.
 \item \lstinline{done}: a list of boolean values with the description of the game status.
 \item \lstinline{info}: a list of \lstinline{Dict} with length of agent number. 
 \end{itemize}
Specifically, each \lstinline{Dict} in \lstinline{info} describes the game information at current step and includes the following items:
\begin{itemize}[leftmargin=20pt]
    \item \lstinline{observation} is a NumPy array including all six components mentioned in Section~\ref{sec:agents}.
    \item \lstinline{legal_action} describes current legal sub-actions with 1 NumPy array. The legal sub-actions incorporates prior knowledge of experienced human players and helps eliminate several unreasonable aspects: 1) skill or attack availability, e.g., the predicted action to release a skill within Cool Down time shall be eliminated; 2) being controlled by enemy hero skill or equipment effects; 3) hero-/item-specific restrictions.
    \item \lstinline{sub_action_mask} is a NumPy array describing dependencies of different \lstinline{Button} actions.
    \item \lstinline{done} describes if the current game is over by a scalar value of 0 or 1. 
    \item \lstinline{frame_no} is the frame number of next state.
    \item \lstinline{reward} is current reward value.
    \item \lstinline{game_id, player_id} are string identifying the game and the run time ID of the current hero.
    \item \lstinline{req_pb} is a string identifying protobuf information of the current hero.
\end{itemize}

We provide an example for using our environment in Listing~\ref{code:example}.

\paragraph{Technical Implementation and  Performance}

\begin{figure}[htb]
\centering    
\vspace{-3mm}
\includegraphics[width=0.6\linewidth]{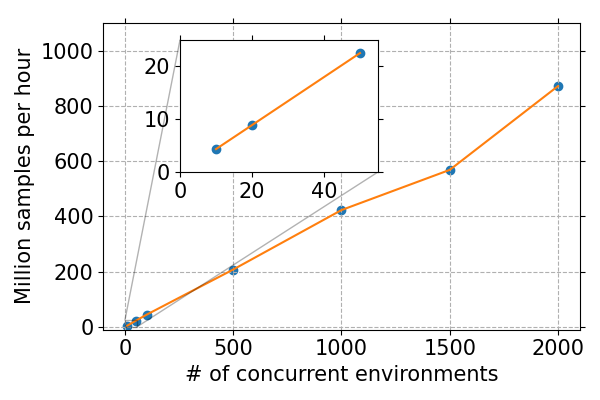}
    \vspace{-3mm}
  \caption{Number of steps per hour versus number of concurrent environments for the \hokenv on a cluster of 10-core Intel Xeon Platinum 8255C CPUs with 2.50GHz. Our infrastructure could generate approximately 4.34 million samples per hour on a single 10-core machine, and could easily scale to larger concurrent environments.}
  \label{fig:infra-perf}
\end{figure}

The \hokenv Engine is written in highly optimized C++ code, allowing it to be run on commodity machines both with GPU and without GPU-based rendering enabled. This allows it to obtain a performance of approximately 4.34 million samples \rebuttal{(around 600 trajectories)} per hour on a single 10-core machine that operates ten concurrent environments (see the subfigure in Figure~\ref{fig:infra-perf}). Furthermore, our infrastructure with sampler, memory pool and learner can easily scale over a cluster of machines. With 200 machines running in total 2000 concurrent environments, our training infrastructure allows over 800 million samples per hour.

\begin{center}
\begin{minipage}{.9\textwidth}

 \vspace{-3mm}
 
\begin{lstlisting}[language=Python, caption=Python example, label=code:example]
from hok import HoK1v1

# load environment
env = HoK1v1.load_game(game_config)

# init agents
agents = []
for i in range(env.num_agents):
    agents.append(Agent(agent_config))
    
# start an episode
obs, _, _, info  = env.reset() 
total_reward = 0.0 
done = False
while not done:
    actions = []
    for i in range(env.num_agents):
        action = agents[i].process(obs[i])
        actions.append(action)
        done = done or info[i]['done']
    obs, reward, done, info = env.step(actions) 

\end{lstlisting}
\end{minipage}
\end{center}

\vspace{-1.5em}
\paragraph{Competition and Evaluation Mechanism} 
Similar to existing game environments, the agent has to interact with the environment and maximize its episodic reward by sequentially choosing suitable actions based on observations of the environment.
The \hokenv supports the competition across different kinds of agents: (1) Rule-based agents. A behavior-tree AI (\bt) is integrated into \hok and provided with the environment. The \bt uses rules that are provided by the game developers and handcrafted differently across different heroes to match the performance of a human player.~\footnote{\rebuttal{In \hok game, human players’ levels are ranked as Bronze, Silver, Gold, Platinum, Diamond, Heavenly, King, and High King (from low to high). BT is hand authored by the game designers to match the Gold level, which can be treated as an entry level for humans in this paper.}} (2) Trained agents. The flexibility on competing against trained agents allows self-play schemes for training RL methods. All the baseline AIs, including the \bt and trained models at different levels, are available. As we will show in Section~\ref{sec:generalization}, these different levels of models are important indicators for researchers to evaluate the performances from different methods. 

\section{Validation}
\label{sec:benchmark}
\vspace{-1em}
The \hokenv is an efficient and flexible learning environment. In order to validate the usefulness of \hokenv, we conduct the evaluation with both heroes as Diaochan for training and implemented two baseline methods. In this scenario, the player needs to destroy one turret before pushing down the opponent's crystal. In the following experiments, unless specified, we keep the heroes to control, the order of item purchasing, and skill upgrading the same for both camps to focus on learning tactics of the agents. 



\vspace{-3mm}
\paragraph{Baselines} In the following experiments, we provide our validation results for existing deep reinforcement learning algorithms, including PPO~\rebuttal{\cite{schulman2017proximal}} and Ape-X DQN~\rebuttal{\cite{dan2019distributed}}. \rebuttal{We acknowledge that establishing baselines for reinforcement learning problems and algorithms could be notoriously difficult, therefore we refer directly to those that proved effective under the Honor of Kings environment~\cite{ye2020mastering,wei2021boosting}.}
For a given algorithm, we ran experiments with similar network architecture, set of hyperparameters, and training configuration as described in the original papers. 


\vspace{-3mm}
\paragraph{Feasibility under different resources}

The first thing we need to validate is whether \hokenv is able to support the training of RL under feasible computing resources. We run the policy over different CPU cores in parallel via self-play with mirrored policies to generate samples. 
We train the PPO network on one NVIDIA Tesla V100 SXM2 GPU. The results are summarized in Table~\ref{tab:resources}. 
Table~\ref{tab:resources} shows that the convergence time with limited resources is less than 7 hours. As the number of CPUs for data collection increases, the training time to beat \bt decreases.
We also conducted experiments on multiple GPUs and found that it was the number of \rebuttal{CPUs} other than \rebuttal{GPUs} that impedes the training time to beat \bt. With more CPU, the environment could generate more training samples, while with more GPU, the training sample could consume more frequently. When the consumption frequency, \ie, number of times per sample used for training, is the same under different $\frac{\#CPUs}{\#GPUs}$, the training time to beat \bt remains approximately the same.

\begin{table}[htb]
\centering
\renewcommand{\arraystretch}{1}

\centering
\vspace{-5mm}
\caption{Training time to beat \bt under different computing resources (standard deviation, each experiment run 3 times). With limited CPU cores, the self-play agent is able to beat \bt using around 6 hours, and the time using additional CPU cores over 512 becomes about one hour.}
\label{tab:resources}

\begin{tabular}{ccc}
\toprule
\begin{tabular}[c]{@{}c@{}} \textbf{CPU Cores}\end{tabular} & \begin{tabular}[c]{@{}c@{}}\textbf{Training hours} \end{tabular} & \begin{tabular}[c]{@{}c@{}}\textbf{Consumption freq.}\end{tabular} \\
 \midrule
128                                                    & 6.16$\pm$0.009                                                                          & 23.80 $\pm$0.87                                \\
256                                                    & 1.67$\pm$0.021                                                                               & 10.60$\pm$0.64                                 \\
512                                                    & 1.08$\pm$0.025                                                                            & 5.21$\pm$0.07                                 \\
1204                                                   & 0.90$\pm$0.06                                                                             & 2.69$\pm$0.03         \\ 
2048                                                   & 0.89$\pm$0.06                                                                             & 1.30$\pm$0.008         \\\bottomrule                                                               
\end{tabular}

\vspace{-5mm}
\end{table}

\begin{figure*}[tbh]
\small
\centering
  \begin{tabular}{ccc}
  \includegraphics[width=.31\linewidth]{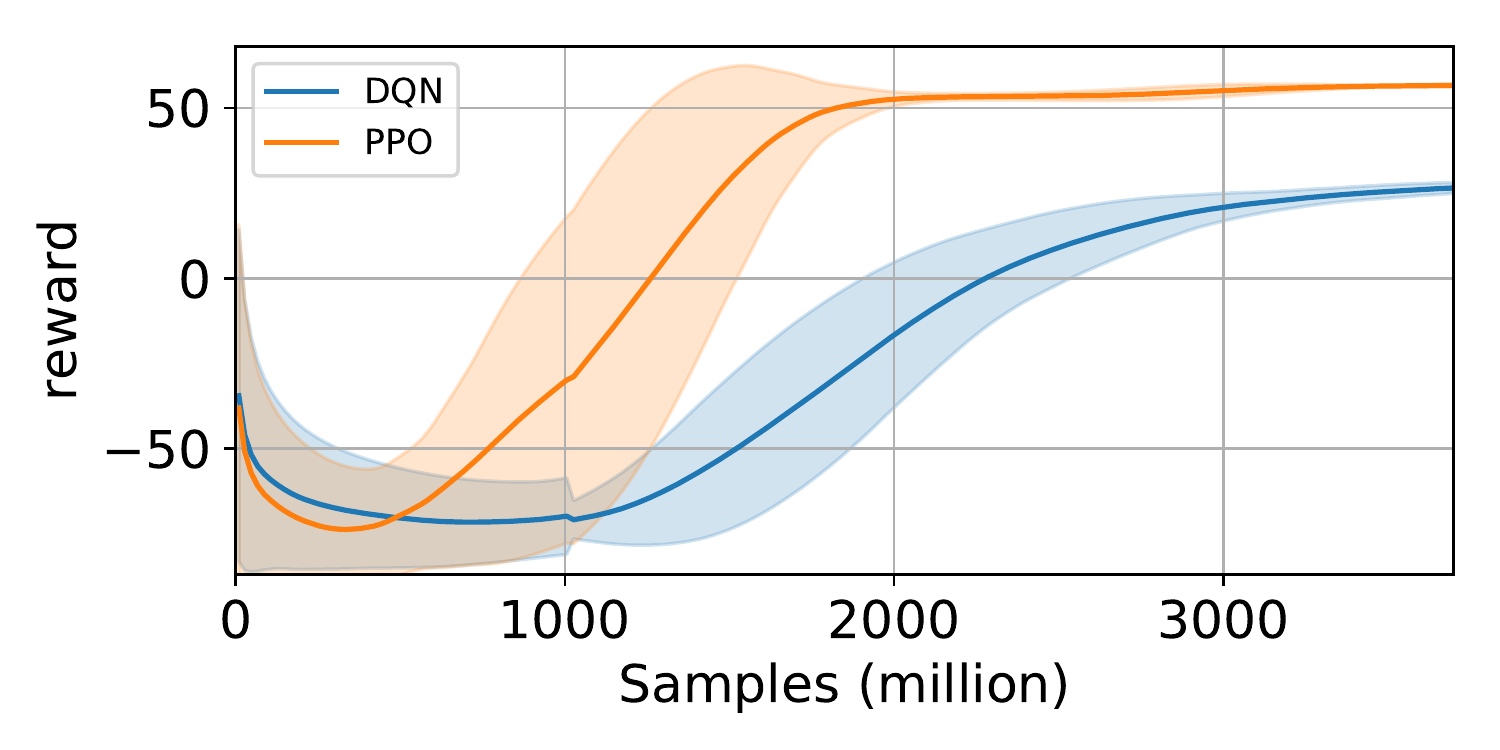} &
  \includegraphics[width=.31\linewidth]{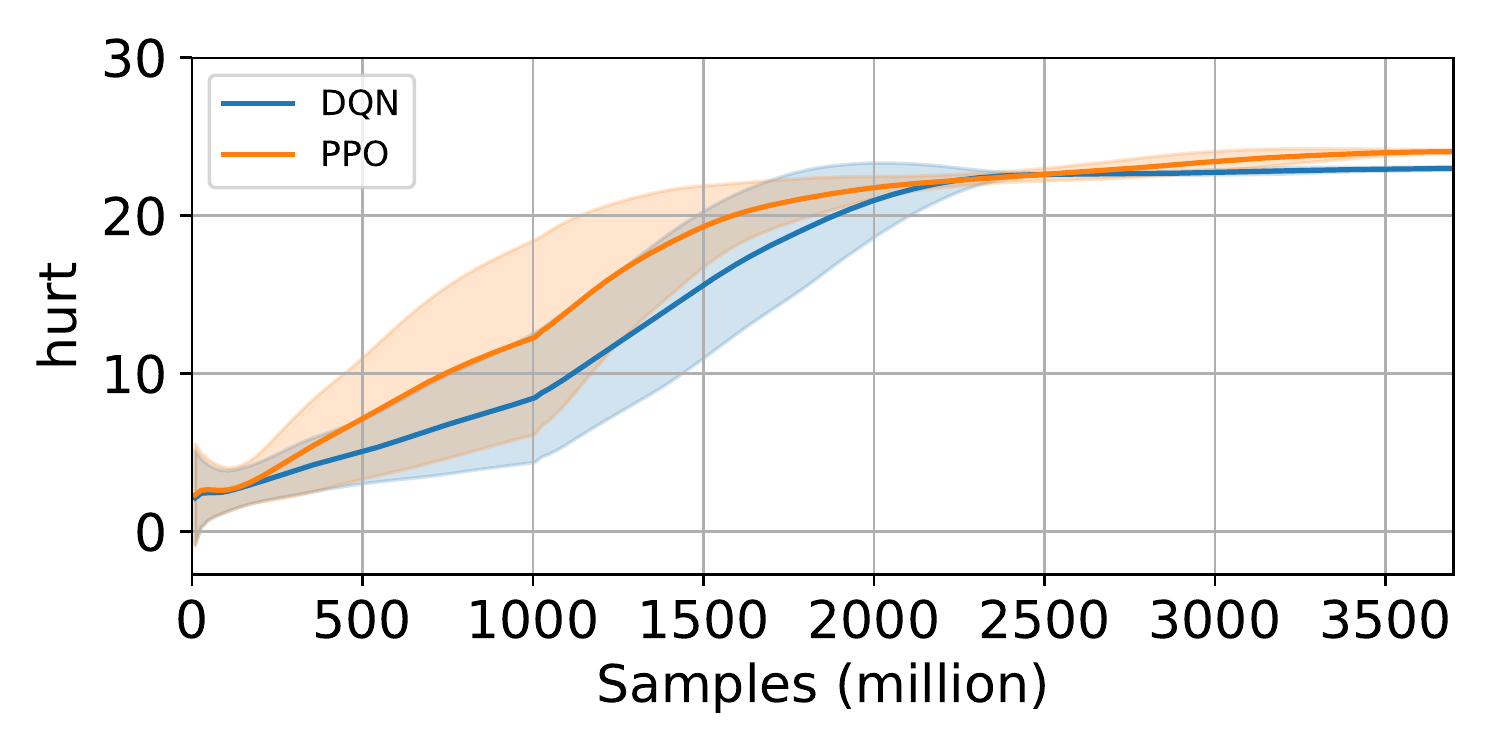} &
  \includegraphics[width=.31\linewidth]{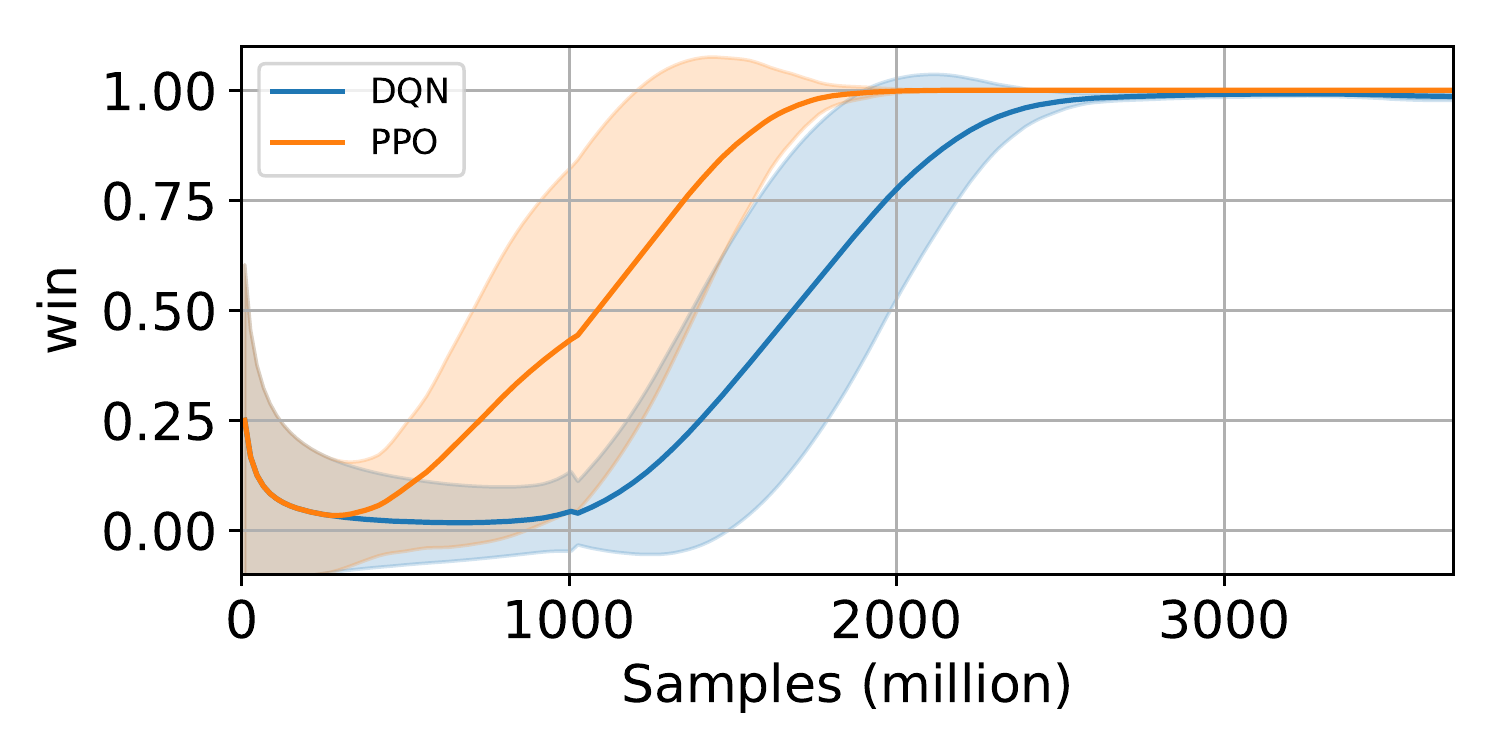}\\
  (a) Reward & (b) Hurt per frame  & (c) Win rate \\
  \end{tabular}
    \vspace{-3mm}
 \caption{Different evaluation metrics on the \hokenv for DQN and PPO w.r.t. the number of training samples. Error bars represent standard deviation. PPO performs better than DQN.}
    \vspace{-5mm}
    \label{fig:baselines}
\end{figure*}

\paragraph{Performance of Different Models}
We provide experimental results for both PPO~\cite{ye2020mastering} and DQN~\cite{wei2021boosting} for the \hokenv in Figure~\ref{fig:baselines}. The experimental results indicate that the \hokenv provides challenges in the choice of RL methods toward a stronger AI, with 2000M samples used for the model of PPO to beat the \bt. Although both PPO and DQN can beat \bt within 3000M samples, they do not achieve the same final performance in terms of reward. 

\paragraph{Performance against \bt} 

\begin{figure}[tbh]
    \vspace{-3mm}
\centering
    \includegraphics[width=0.5\linewidth]{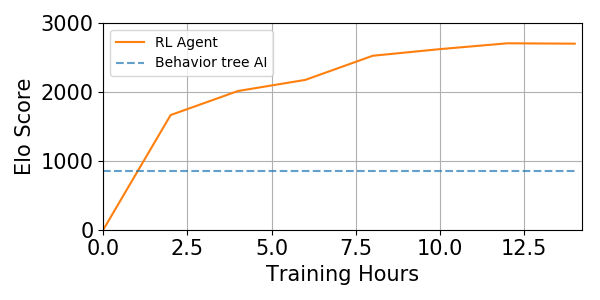}
    \vspace{-3mm}
  \caption{The Elo score of models under different training times. Blue lines shows the Elo score for \bt. The self-play trained PPO could beat \bt in about 1 hour, and converge after 10 hours.}
    \vspace{-5mm}
  \label{fig:elo-score}
\end{figure}

We present preliminary results showing the possibility of using RL to achieve human-level performance. We run experiments with over 1024 CPU cores and one GPU core via self-play. 
The Elo score~\cite{elo2008} is used to evaluate the relative performance of one model among all the models.
From Figure~\ref{fig:elo-score}, we can see that PPO can beat \bt (indicating the level of a normal human), within two hours and takes about 10 hours to converge. 

\section{Research Direction: Generalization Challenge in Competitive RL}
\label{sec:generalization}
In \hokenv, players could control different heroes against different opponents, and the environment is also changed by the hero's actions. A policy with good transferability is expected to perform stably against different opponents. We conducted two experiments with PPO algorithm to showcase the transferability challenge.



\begin{figure}[htb]
\centering
  \includegraphics[width=0.98\linewidth]{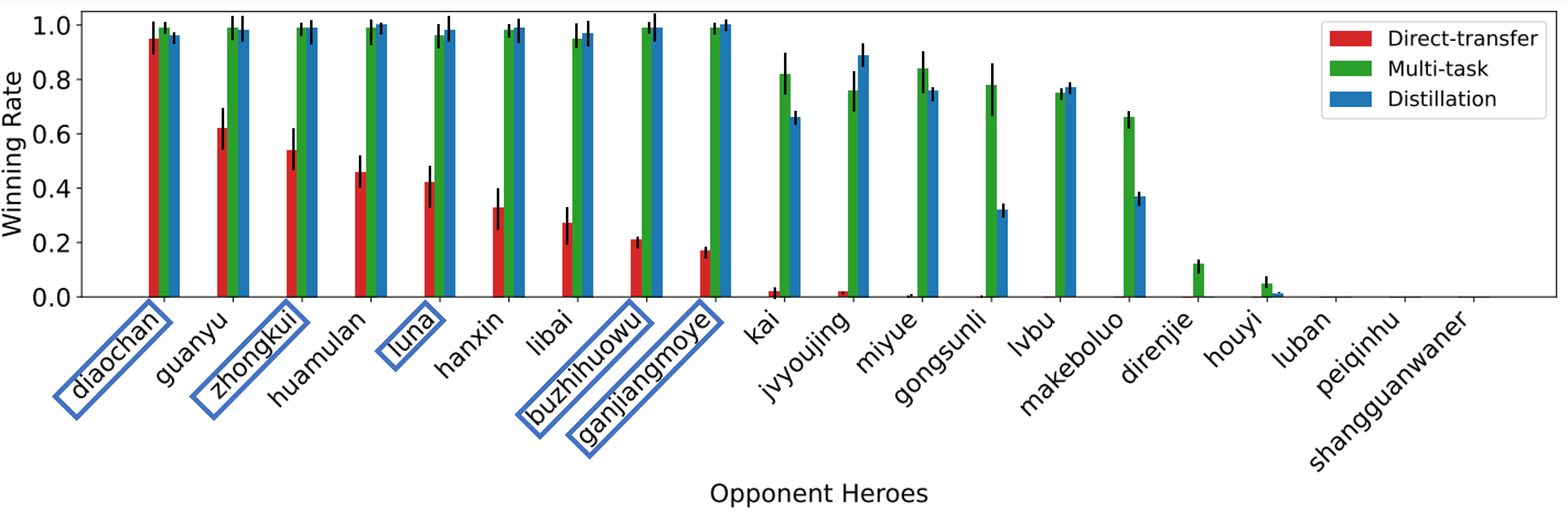}
  \vspace{-3mm}
  \caption{Win rate of a well-trained model from task "Diaochan (RL) vs. 
  Diaochan (\bt)" transferred to tasks "Diaochan (RL) vs. different opponent heroes (\bt)". The agent is trained to control Diaochan against Diaochan controlled by \bt, and tested to control Diaochan against different heroes controlled by \bt.  Red: Directly transferring the model to control Diaochan and compete with different opponent heroes. Green: Multi-task training on five tasks "Diaochan (RL) vs. Diaochan/Buzhihuowu/Luna/Ganjiangmoye/Zhongkui (\bt)" and testing the model on twenty tasks. Blue: Distilling the model trained from five tasks "Diaochan (RL) vs. Diaochan/Buzhihuowu/Luna/Ganjiangmoye/Zhongkui (\bt)" and testing the model on twenty tasks. The policy trained on Diaochan could not generalize to all tasks with different \emph{opponent} heroes. \rebuttal{Blue rectangles highlights the five tasks used in multi-task and distillation. The error bars indicate the standard deviation under five seeds.}}
  \label{fig:challenge-oppo}
  \vspace{-2mm}
\end{figure}

\vspace{-3mm}
 \paragraph{Generalization Challenge across Opponents} In the first experiment, the policy is trained under the task of "Diaochan (RL) vs. Diaochan (\bt)", and tested under the tasks of "Diaochan (RL) vs. different opponent heroes (\bt)" for 98 rounds of matches. In these tasks, the RL model is only used to control Diaochan, and its opponents are changed as different tasks. The results are shown in Figure~\ref{fig:challenge-oppo}. The model trained on Diaochan could beat the opponent Diaochan controlled by \bt (90\% winning rate) since the target to control during testing and training is the same. However, when the opponent hero changes, the performance of the same trained policy drops dramatically, as the change of opponent hero differs the testing setting from the training setting, indicating the lack of transferability of the policy learned by existing methods. 

\begin{figure}[t!]
\centering
  \includegraphics[width=0.98\linewidth]{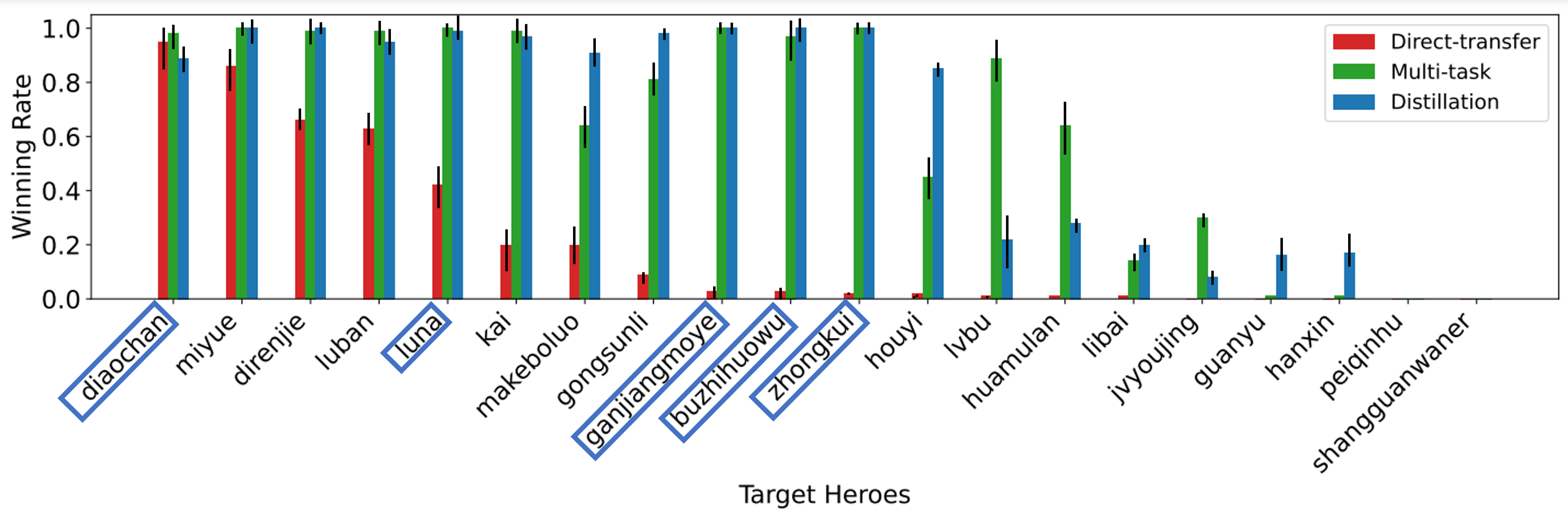}
  \caption{Win rate of a well-trained model from task "Diaochan (RL) vs. 
  Diaochan (\bt)" transferred to tasks "Different target heroes (RL) vs. Diaochan (\bt)". The agent is trained to control Diaochan against Diaochan controlled by \bt, and tested to control different heroes against Diaochan controlled by \bt. Red: Directly transferring the model to control Diaochan and compete with different opponent heroes. Green: Multi-task training on five tasks "Diaochan/Buzhihuowu/Luna/Ganjiangmoye/Zhongkui (RL) vs. Diaochan (\bt)" and testing the model on twenty tasks. Blue: Distilling the model trained from five tasks "Diaochan/Buzhihuowu/Luna/Ganjiangmoye/Zhongkui (RL) vs. Diaochan (\bt)" and testing the model on twenty tasks. The policy trained on Diaochan could not generalize to all tasks with different \emph{target} heroes. \rebuttal{Blue rectangles highlights the five tasks used in multi-task and distillation. The error bars indicate the standard deviation under five seeds.}}
  \label{fig:challenge-targets}
    \vspace{-5mm}
\end{figure}

\vspace{-3mm}
 \paragraph{Generalization Challenge across Targets} In the second experiment, the policy is trained under the task of "Diaochan (RL) vs. Diaochan (\bt)", and tested under the task of "different target heroes (RL) vs. Diaochan (\bt)" for 98 rounds of matches. In these tasks, the RL model is used to control different heroes, but their opponents remains the same as Diaochan. The results are shown in Figure~\ref{fig:challenge-targets}. When the target changes from Diaochan to other heroes, the performance of the same trained policy drops dramatically, as the change of target heroes differs the action meaning from the Diaochan's action in the training setting.

 \paragraph{Remedies: Multi-task and Distillation} A possible fix to enhance transferability in RL policies is to do multi-task training, \ie, adding multiple testing settings during training, and forcing the training setting the same as the testing. We conducted an additional experiment to show that these methods show improvements on certain tasks. For multi-task models, we train the model with five tasks "Diaochan (RL) vs. Diaochan/Buzhihuowu/Luna/Ganjiangmoye/Zhongkui", and test it in twenty tasks.  The results are shown in both Figure~\ref{fig:challenge-oppo} and Figure~\ref{fig:challenge-targets}. We can see that multi-task training improves the performance in all the test tasks. 
 
 
 Another possible enhancement for transferability is to distill one model from multiple models. We use the student-driven policy distillation that is proposed in \cite{czarnecki2019distilling}. We conducted an experiment to show that this method could achieve similar improvements to multi-task training over directly transferring models. In both Figure~\ref{fig:challenge-oppo} and Figure~\ref{fig:challenge-targets}, the model is distilled from five models trained on five different tasks and tested in twenty tasks. We can see that the distillation improves the performance in all the test tasks. 
 
 In the experiments above, we only chose the task "Diaochan vs. Diaochan" as the primary training task. However, other feasible training tasks, such as "Buzhihuowu vs. Buzhihuowu", are reasonable options. We chose 5 different heroes as the primary training tasks and test generalization across both opponents and targets with direct transfer in Appendix~\ref{append:results}.
 
 \paragraph{Multi-level Models for Evaluation} Although multi-task and distillation could improve the generalization performance on many tasks, we also noticed that for certain tasks like opponent heroes changed to Peiqinhu/Shangguanwaner in Figure~\ref{fig:challenge-oppo}, or target heroes changed to Peiqinhu/Shangguanwaner in Figure~\ref{fig:challenge-targets}, the winning rate for the Diaochan model is constantly zero. This makes it hard to evaluate the performance of different techniques for generalization. 
 
\begin{figure}[htb]
\centering
    \includegraphics[width=0.75\linewidth]{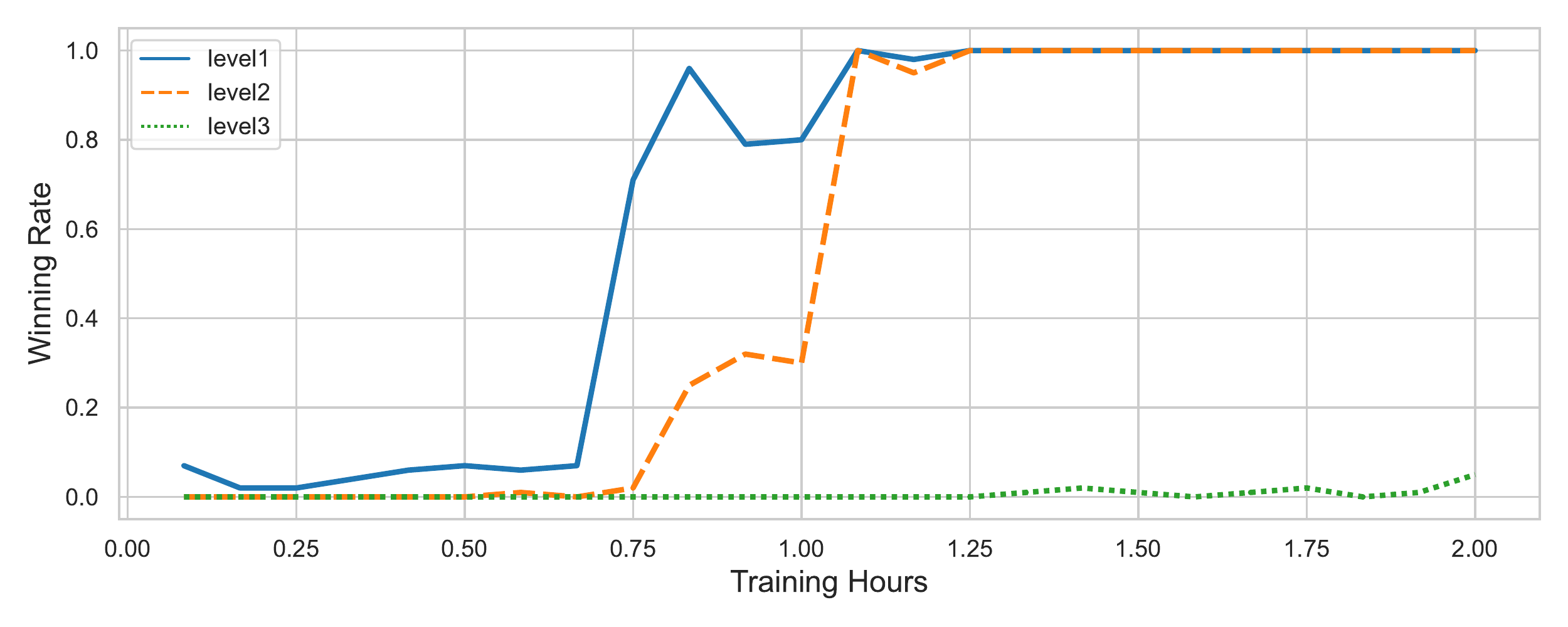}
    \vspace{-3mm}
  \caption{Winning rate of an RL agent for Diaochan to beat an opponent hero Buzhihuowu at different levels. }
  \label{fig:level}
\end{figure}

 Similarly, we noticed that both multi-task and distallation achieves near 100\% winning rate in tasks like opponent heroes changed to Buzhihuowu \etc in Figure~\ref{fig:challenge-oppo}, or target heroes changed to Buzhihuowu \etc in Figure~\ref{fig:challenge-targets}. To help researchers better evaluate their models, we also provided different levels of models for different heroes other than \bt. One example of different levels of Buzhihuowu as the opponents for Diaochan is shown in Figure~\ref{fig:level}. If we only provide the level-3 model of Buzhihuowu, the winning rate would be tricky and hard to tell the actual performance of trained Diaochan model in different training hours.
 
\section{Conclusion and Future Work}
\label{sec:conclusion}
The \hokenv is a starting place for the design and performance comparison of competitive reinforcement learning algorithms with generalization necessity across different tasks. The configurable state and reward functions allow for easy-to-use suite-wide performance measures and explorations. The results presented in this paper constitute baselines using well-performing implementations of these algorithms. 

We are excited to share the \hokenv with the broader community and hope that it will be helpful. We look forward to the diverse research the environment may enable and integrating community contributions in future releases. In the future, we plan to optimize the deployment of our environment across different platforms. Meanwhile, we will also organize more competitions based on \hokenv and provide our computing resources to contestants. 

\section*{Acknowledgement}
The SJTU team is partially supported by ``New Generation of AI 2030'' Major Project (2018AAA0100900) and National Natural Science Foundation of China (62076161).
The author Jingxiao Chen is supported by Wu Wen Jun Honorary Doctoral Scholarship, AI Institute, Shanghai Jiao Tong University.




\newpage
\appendix

\section{License and Documentations}
\label{supp:doc}
\hokenv is open-sourced under Apache License V2.0. 
The code for agent training and evaluation is built with official authorization from \hok and is available at: ~\url{https://github.com/tencent-ailab/hok_env}, containing the game engine and the affiliated code for agent training and evaluation. 
The encrypted game engine and game replay tools can be downloaded from: \url{https://aiarena.tencent.com/hok/download}. 
All experiments can be reproduced from the source code, which includes all hyper-parameters and configuration.
\hokenv is authorized by the game developers of \hok and the authors will bear all responsibility in case of violation of rights, etc., ensure access to the data and provide the necessary maintenance.
We also provide the detailed documentations for \hokenv and \hok in this link: \url{https://aiarena.tencent.com/hok/doc/}, including the following:

\begin{itemize}
    \item Quick start of \hokenv
        \begin{itemize}
        \item Installation
        \item First agent
        \item Evaluation
    \end{itemize}
    \item Description of \hokenv
    \begin{itemize}
        \item Game Description of \hok
        \item Observation space
        \item Action space
        \item Reward
    \end{itemize}
    \item Game Description of \hok
    \item \hokenv package API Reference
\end{itemize}

\section{Basic Units}
\label{append:basic}

\begin{figure}[hbt]
  \centering
  \vspace{5mm}
   \includegraphics[width=\linewidth]{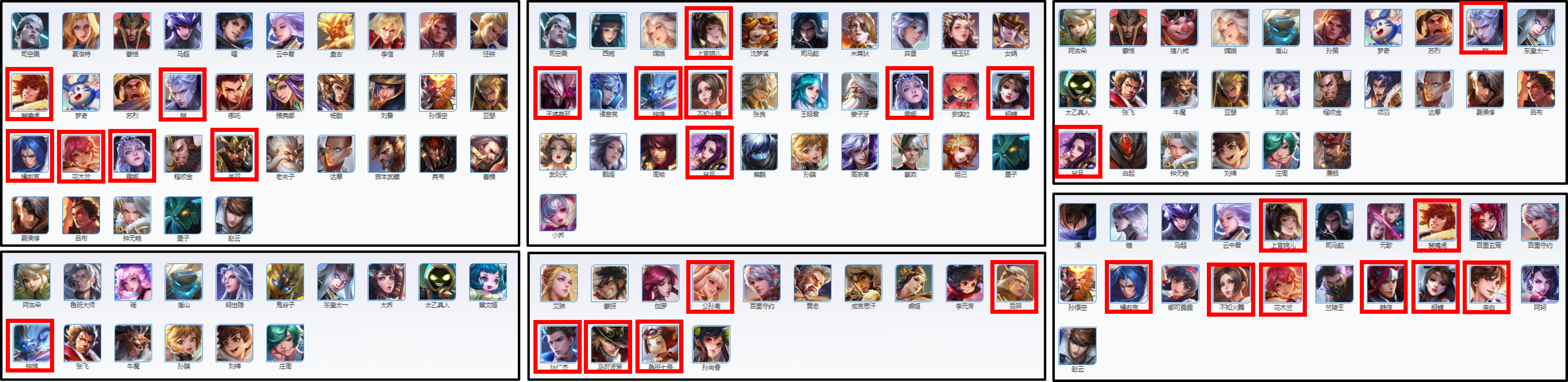} 
 \caption{Subset of the \hok heroes. There are six general types of heroes and each hero has a specific set of hero attributes and skills. Red rectangles highlight the heroes provided in current~\hokenv.}
    \label{fig:hero}
\end{figure}

\paragraph{Hero}  Heroes are player-controlled units that can move around and have abilities to release various attacking and healing skills. There are many heroes with different attributes, skills and tactics in the full game settings for \hok. Each player can choose one hero to control and team up with others as a lineup of heroes. For more information on hero skills, please refer to the official website of \textit{Honor of Kings}: \url{https://pvp.qq.com/m/m201706/heroList.shtml}. As is shown in Figure~\ref{fig:hero}, \hokenv now supports 20 heroes, because these 20 heroes are sufficient for testing the generalization ability. We will support more heroes in the future.

\paragraph{Creep} Creeps are a small group of computer-controlled creatures that periodically travel along the predefined lane to attack the opponent units.
\paragraph{Turret} Turrets, i.e., defensive buildings, are designed to attack any opponent units moving into their sight area. 
\paragraph{Crystal} The crystal, located in the base of each player, can attack opponent units around their sight area with higher damages than turrets. The final victory condition is to push down the enemy's crystal. And the opponent units cannot cause damage to the base towers without destroying the turrets.

\section{Hero Details}
\label{append:agent}

\subsection{Basic Attributes}
The basic attributes are what every hero has: health point (HP) volume and magical points (MP) volume, attack, defense, resistance, \etc.

\begin{table}[hbt]
\centering
\caption{Basic attributes of a hero}
\begin{tabular}{cc}
\toprule
Maximum HP            &  Maximum MP  \\
Physical attack       &  Magical attack     \\
Physical bloodsucking &  Magical bloodsucking \\
Physical defense      &  Magical defense\\
Physical penetration  &  Magical penetration\\
HP regeneration speed  &  MP regeneration speed\\
Attack speed          &  Cool-down time \\
Bonus attack chance  &  Resilience  \\
Attack range    & Movement speed \\
\bottomrule
\end{tabular}
\end{table}

Among all the basic attributes, HP and MP are essential for the states of the hero, where the hero would die when HP runs out and the hero would not be able to release magical skills if MP runs out.

Defense and attack exist in pairs. There are three types of attacks: physical damage, magic damage, and true damage. Among them, physical defense is only effective for physical damage, magical defense is only effective for magical damage, and true damage is effective regardless of physical and magical defense.

\begin{figure}[tbh]
\centering
    \includegraphics[width=0.5\linewidth]{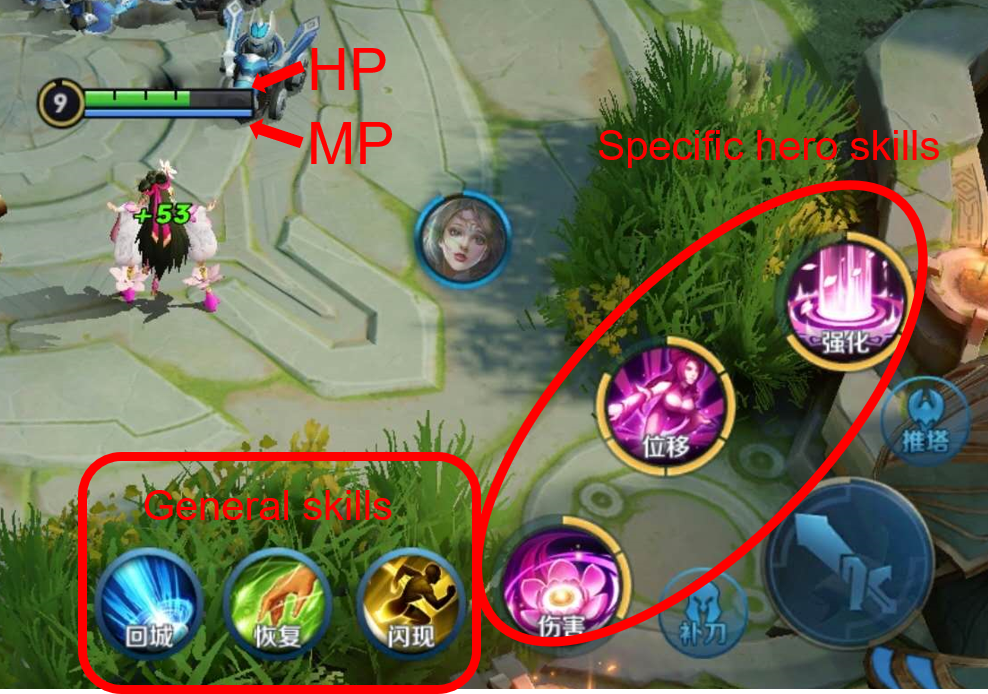}
  \caption{Important basic attributes of the hero like HP and MP are shown in the UI of \hok, together with the state of hero skills.}
\end{figure}

\subsection{Skills}
\paragraph{General skills}
General skills are independent skills players need to choose according to their hero type before the game starts. The general skill does not require any consumption of experiences or gold. Once used by the hero,  it will be available only after a certain cooling-down time.

\paragraph{Specific hero skills} 
Each hero has a specific set of hero skills, which is different from all other heroes. Each skill has different levels, which can be upgraded with experiences from killing or clearing opponent units. It usually consumes MP to release specific hero skills.

\subsection{Hero Types}
Generally, there are six types of heroes with different playing strategies, including:
\begin{itemize}
    \item Tank. Tank-type heroes have high blood and defense, and are heroes with strong survivability and average damage ability. They usually occupy the front row position in team battles, resist the damages from enemies.
    \item Warrior. Warrior-type heroes have balanced offensive and defensive capabilities. They usually stand behind tank heroes in team battles, take a small amount of enemy damage, fight in the enemy lineup, and sometimes act as a vanguard in the absence of tank heroes and take a lot of damage.
    \item Assassin. Assassin-type heroes are heroes with weak survivability, but extremely explosive damage capabilities. In a team battle, the assassin can go to the back of the enemy lineup and find the right time for a set of skills to kill the enemy heroes with low blood.
    \item Mage. Mage-type heroes are heroes with weak survivability, but high magical damage and control skills. They usually occupy the back-row position in the team battle and try to control the enemy heroes.
    \item Marksman. Marksman-type heroes are heroes with extremely high remote physical damage capabilities and control skills. In team battles, Marksman can use their remote damage advantage to stand in the back row.
    \item Support. Support heroes are heroes with mediocre survivability and output ability. They rely on powerful skills to increase the status of teammates and control enemy heroes.
\end{itemize}

\subsection{Items}
In \hok, heroes can not only upgrade to higher-level skills with experiences but also can purchase items with golds. Each hero is allowed to possess up to six items. The types of items in \hok are various and complex. When choosing items, the players need to decide which items to buy, keep and sell to strengthen the hero's characteristics. For example, tank-type heroes might need shields to enhance defense, warriors and mages might need to enhance their damage, \etc. However, the choices of items are also influenced by the opponents and their items during different phases of the game. For example, Suppose the opponent is a mage that relies on magical attacks. In that case, the player can choose magical resistance items, making it difficult for the opponent to do more magical damage.


\section{State Vector}
\label{append:herostate}
Users can customize and redefine their own observations from the returned `info' from `env.step()', while we provide a basic set of observations here.
The observation space of \hokenv consists of five main components, whose dimensions depend on the number of heroes in the game:
The most notable component is \lstinline|HeroStatePublic|,  which describes the hero's status, including whether it's alive, its ID, its health points (HP), and its skill status, and its skill related information, \etc. \lstinline|HeroStatePrivate| includes the specific kill information for all the heroes in the game. \lstinline|VecCreeps| describes the status of soldiers in the troops, including their locations, HPs, \etc. \lstinline|VecTurrets| describes the status of turrets and crystals. It should be noted that the states of enemy units, even if they're invisible to the ego camp, are also available, which could be helpful for algorithms that learn from hindsight and should be masked out during policy execution. The last part of the observation space is \lstinline|VecCampsWholeInfo|, which indicates the period of the match.
Their dimensions and descriptions can be found in Table~\ref{supp:state}. The meanings of every dimension in the state space can be found in the link provided in Documentation of Section~\ref{supp:doc}. 
\begin{table*}[htb]
\centering
\small
\caption{State information}
\label{supp:state}
\begin{tabular}{ccp{6cm}c}
\toprule
Feature Class                                                                      & Field              & Description                                                                                                                                                 & Dimension \\ \midrule
\multirow{2}{*}{\lstinline|HeroStatePublic|} & Hero Status & HP, MP, level, exp, position, skills, \etc & 49 * 2      \\ 
                         & Hero skills & Specific attributes of hero skills, like buff mark layers, \etc & 53 * 2 \\ \midrule
\multirow{4}{*}{\lstinline|HeroStatePrivate|} & Diaochan-related & skill position and buffs & 11      \\ 
                         & Luna-related & phases and status of normal attack & 7 \\
                         & Jvyoujing-related & strengthened normal attack related & 9 \\ 
                         & Luban-related & attack stage & 5 \\ \midrule 
                         
\multirow{2}{*}{\lstinline|VecCreeps|} & Status & HP, camp, attack range, etc. & 12 * 4      \\ 
                         & Position & Absolute and relative coordinates, distance & 6 * 4 \\ \midrule  
\multirow{2}{*}{\lstinline|VecTurrets|} & Status & Hp, camp, attack range, etc. & 12 * 4      \\ 
                         & Position & Absolute and relative coordinates, distance & 6 * 4 \\ \midrule 
\lstinline|VecCampsWholeInfo|                         & Current Period             & Divide game time into 5 periods     & 5       \\ \midrule
Total                              & Sum             & All features     & 491 \\ \bottomrule
\end{tabular}
\end{table*}


\section{Action Space}
\label{append:actionspace}

\begin{figure}[tbh]
\centering
 \includegraphics[width=0.5\linewidth]{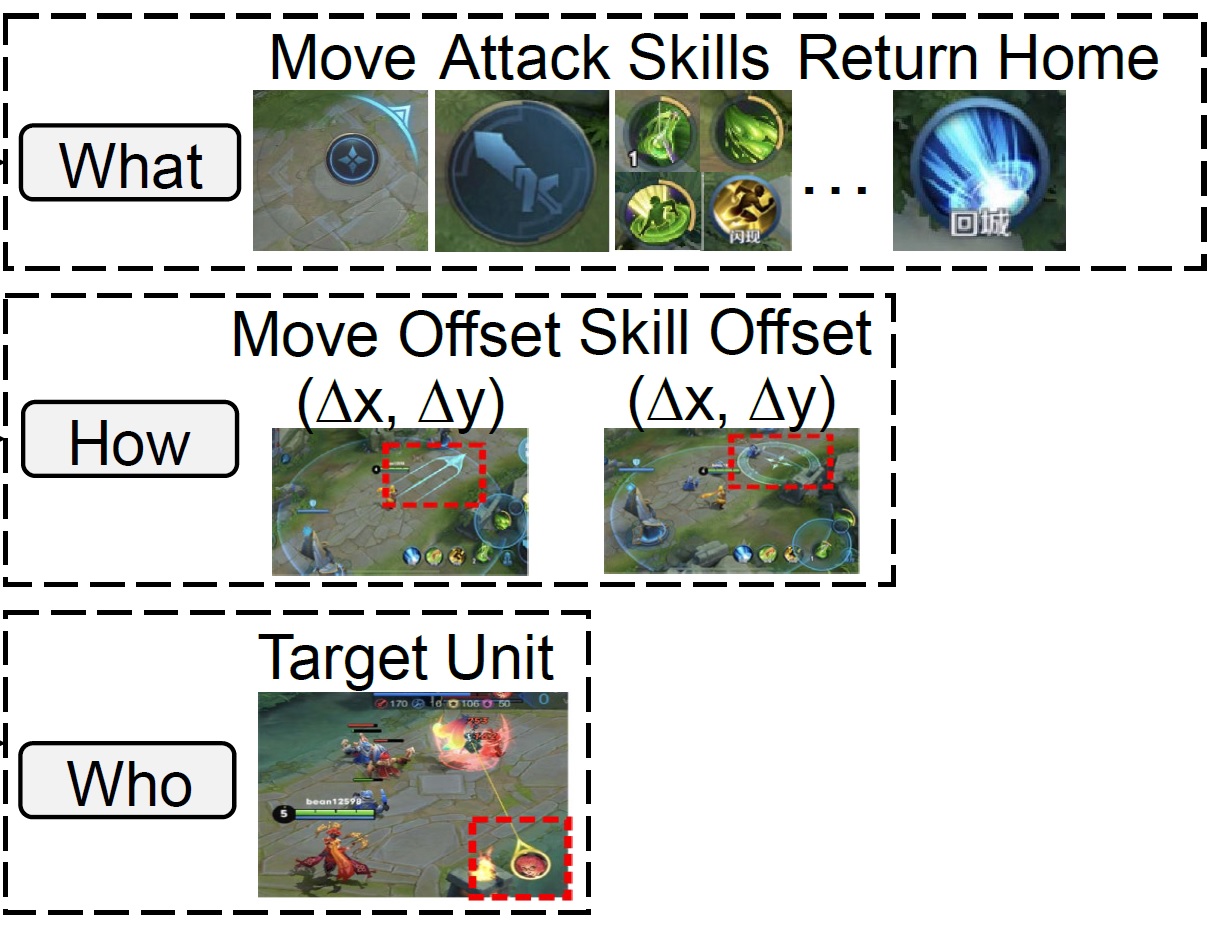}
  \caption{The action space is composed of action button, offsets and targets.}
\label{supp:fig:action}
\end{figure}

The native action space of the environment consists of a triplet form, \ie, the action button, the movement offset over the x-axis and y-axis, the skill offset over the x-axis and y-axis, and the target game unit. It covers all the possible actions of the hero hierarchically: 1) what action button to take; 2) who to target, e.g., a turret, an enemy hero, or a soldier in the troop; 3) how to act, e.g., the discretized direction to move and release skills. As illustrated in Figure~\ref{supp:fig:action}, the action buttons include move, attack, skill releasing, etc. The details of the action space, including the types and meanings of every dimension, can be found in Table~\ref{supp:action}.

\begin{table*}[t!]
\centering
\small
\caption{Action Space description}
\label{supp:action}
\begin{tabular}{cccc}
\toprule
Action Class                & Type      & Description       & Dimension     \\ \midrule

\multirow{11}{*}{Button}    & None          & No action                 & 1             \\ 
                            & Move          & Move hero                 & 1             \\ 
                            & Normal Attack & Release normal attack     & 1             \\ 
                            & Skill 1        & Release skill 1st         & 1             \\
                            & Skill 2        & Release skill 2nd         & 1             \\
                            & Skill 3        & Release skill 3rd         & 1             \\ 
                            & Heal Skill    & Release heal skill        & 1             \\
                            & Summoner Skill & Release summoner skill  & 1              \\ 
                            & Recall        & \begin{tabular}[c]{@{}c@{}}Start channeling and return to home crystal \\ after a few seconds if not interrupted\end{tabular}  & 1             \\ 
                            & Skill 4        & Release skill 4th (Only valid for certain heroes) & 1 \\
                            & Equipment Skill & Release skill provided by certain equipment & 1 \\ \midrule
\multirow{2}{*}{Move}                        & Move X        & Move direction along X-axis      & 16            \\
                            & Move Z        & Move direction along Z-axis      & 16            \\ \midrule
\multirow{2}{*}{Skill}                       & Skill X        & Skill direction along X-axis      & 16            \\
                            & Skill Z        & Skill direction along Z-axis      & 16            \\ \midrule
\multirow{5}{*}{Target}     & None      & Empty target      & 1             \\ 
                            & Self      & Self player       & 1             \\ 
                            & Enemy     & Enemy player      & 1             \\ 
                            & Soldier   & 4 Nearest soldiers & 4             \\ 
                            & Tower     & Nearest tower     & 1             \\ \bottomrule
\end{tabular}
\end{table*}

\section{Reward Information}
\label{append:rewarddesign}
Users can customize and redefine their own rewards (including the termination rewards) from the returned `info' from `env.step()', while we provide a basic set of rewards here.
\hok has both sparse and dense reward configurations in five categories: 1) Farming related: the amount of gold and experience, and the penalty of not acting, which are dense reward signals; 2) KDA related: the number of kill, death and assist, and the last hit to enemy units, which are sparse reward signals; 3) Damage related: a dense reward - the number of health point, and a sparse reward -  the amount of attack to enemy hero; 4) Pushing related: the amount of attack to enemy turrets and crystal, which are dense rewards; 5) Win/lose related: destroy the enemy home base, which are sparse reward signals received at the end of the game.

It is possible that only using the dense reward version of the \hok will likely resemble the sparsity, which is often seen in previously sparse rewarding benchmarks. Given the sparse-reward nature of this task, we encourage researchers to develop novel intrinsic reward-based systems, such as curiosity, empowerment, or other signals to augment the external reward signal provided by the environment.

\begin{table}[htb]
\small
\caption{Reward Design}
\centering
\begin{tabular}{cccp{5cm}}
\toprule
Reward                  & Weight    & Type      & Description \\ \midrule
hp\_point               & 2.0       & dense     & the rate of health point of hero \\ 
tower\_hp\_point        & 10.0      & dense     & the rate of health point of tower \\ 
money (gold)            & 0.006     & dense     & the total gold gained \\ 
ep\_rate                & 0.75      & dense     & the rate of mana point \\ 
death                   & -1.0      & sparse    & being killed \\ 
kill                    & -0.6      & sparse    & killing an enemy hero \\ 
exp                     & 0.006     & dense     & the experience gained \\ \bottomrule
\end{tabular}
\end{table}






\section{Hyperparameters}
\label{append:hyper}

The network structure is directly adopted from~\cite{ye2020mastering} whose implementation can be found in our code~\url{https://github.com/tencent-ailab/hok_env/blob/main/code/common/algorithm.py}. We use Adam optimizer with initial learning rate 0.0001.
For PPO, the two clipping hyperparameters $\epsilon$ and $c$ are set as 0.2 and 3, respectively. The discount factor is set as 0.997. For the case of Honor of Kings, this discount is valuing future rewards with a half-life of about 46 seconds. We set $\lambda = 0.95$ in GAE to reduce the variance caused by delayed effects. For DQN, the discount factor of tagert Q-network is set as 0.98.

\section{Additional Experiments}
\label{append:results}

The \hokenv provides scenarios and flexible APIs to train different reinforcement learning models. When training deep RL models, it often requires crafting designs like network structure and loss function based on different problems. \rebuttal{In the following sections, if not specified, we will use Diaochan vs. Diaochan as our task in the experiment.}

\subsection{Dual clip in PPO} 
\begin{figure*}[tbh]
\centering
  \begin{tabular}{ccc}
  \includegraphics[width=.31\linewidth]{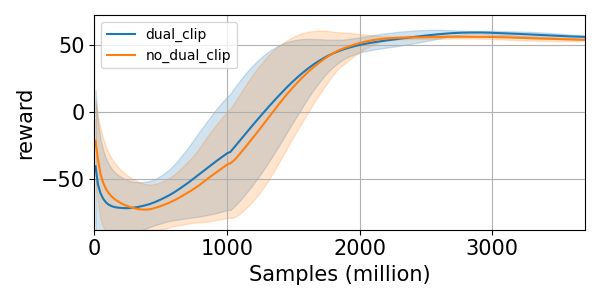} &
  \includegraphics[width=.31\linewidth]{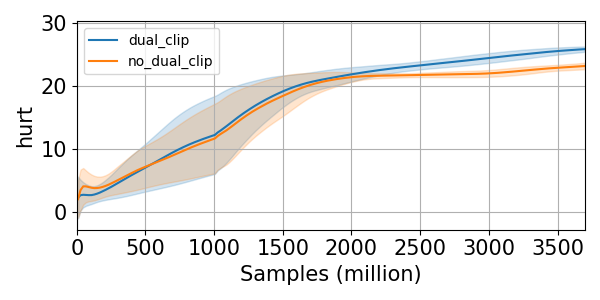} &
  \includegraphics[width=.31\linewidth]{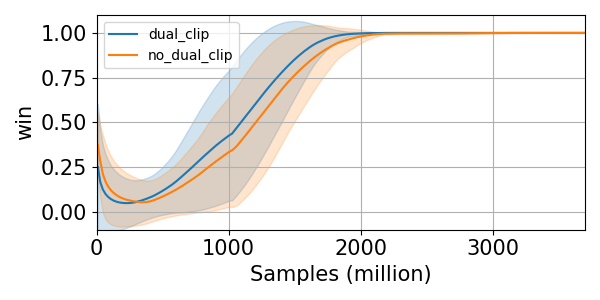}\\
  (a) Reward & (b) Hurt per frame  & (c) Win rate \\
  \end{tabular}
 \caption{Different evaluation metrics on the \hokenv 1v1 mode for the ablation study on dual-clip w.r.t. the number of training samples. Error bars represent standard deviation.}
    \label{fig:clip}
\end{figure*}

To tackle the dramatic shift between different rounds of data in off-policy reinforcement learning, one common design to apply clipping, on either policy entropy or trajectory importance ratios~\cite{schulman2017proximal}. In~\cite{ye2020towards}, it utilized a dual-clip PPO to support large-scale distributed training. And we investigate the influences of this design and found out the model with dual clip is slightly better than the original PPO model, as is shown in Figure~\ref{fig:clip}.

\subsection{Legal action mask} 
\begin{figure*}[htb]
\centering
  \begin{tabular}{ccc}
  \includegraphics[width=.31\linewidth]{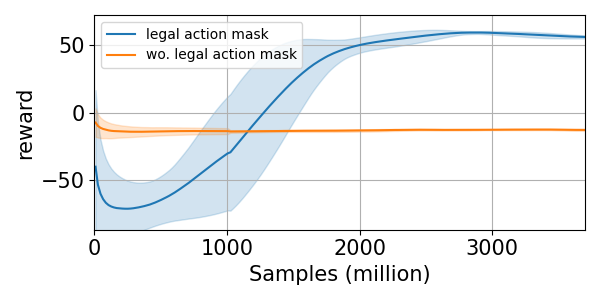} &
  \includegraphics[width=.31\linewidth]{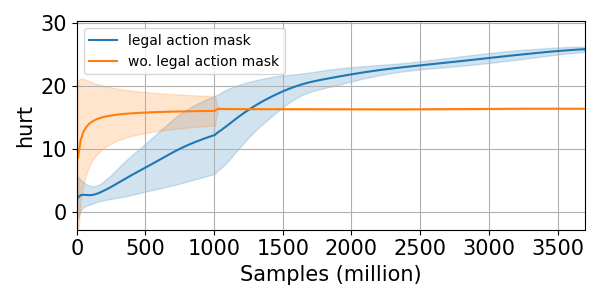} &
  \includegraphics[width=.31\linewidth]{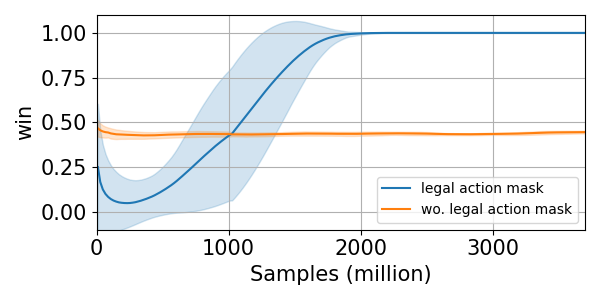}\\
  (a) Reward & (b) Hurt per frame  & c) Win rate \\
  \end{tabular}
 \caption{Different evaluation metrics on the \hokenv 1v1 mode for the ablation study on legal action mask. Error bars represent standard deviation.}
    \label{fig:legal}
\end{figure*}

To improve the training efficiency and incorporate the prior knowledge of human players, especially for RL problems with large action spaces. In \hokenv, it provides legal action information to help eliminate unreasonable actions for reference, where researchers can design their own legal actions based on their knowledge. We can use the legal action information to mask unreasonable actions out during the training process. The experiment results is shown in Figure~\ref{fig:legal}. As expected, without legal action, the agent quickly converges to a local optimum under the large action space, and it is critical to use action mask for better performances.


\subsection{LSTM for partial observation} 
\begin{figure*}[t!]
\centering
  \begin{tabular}{ccc}
  \includegraphics[width=.31\linewidth]{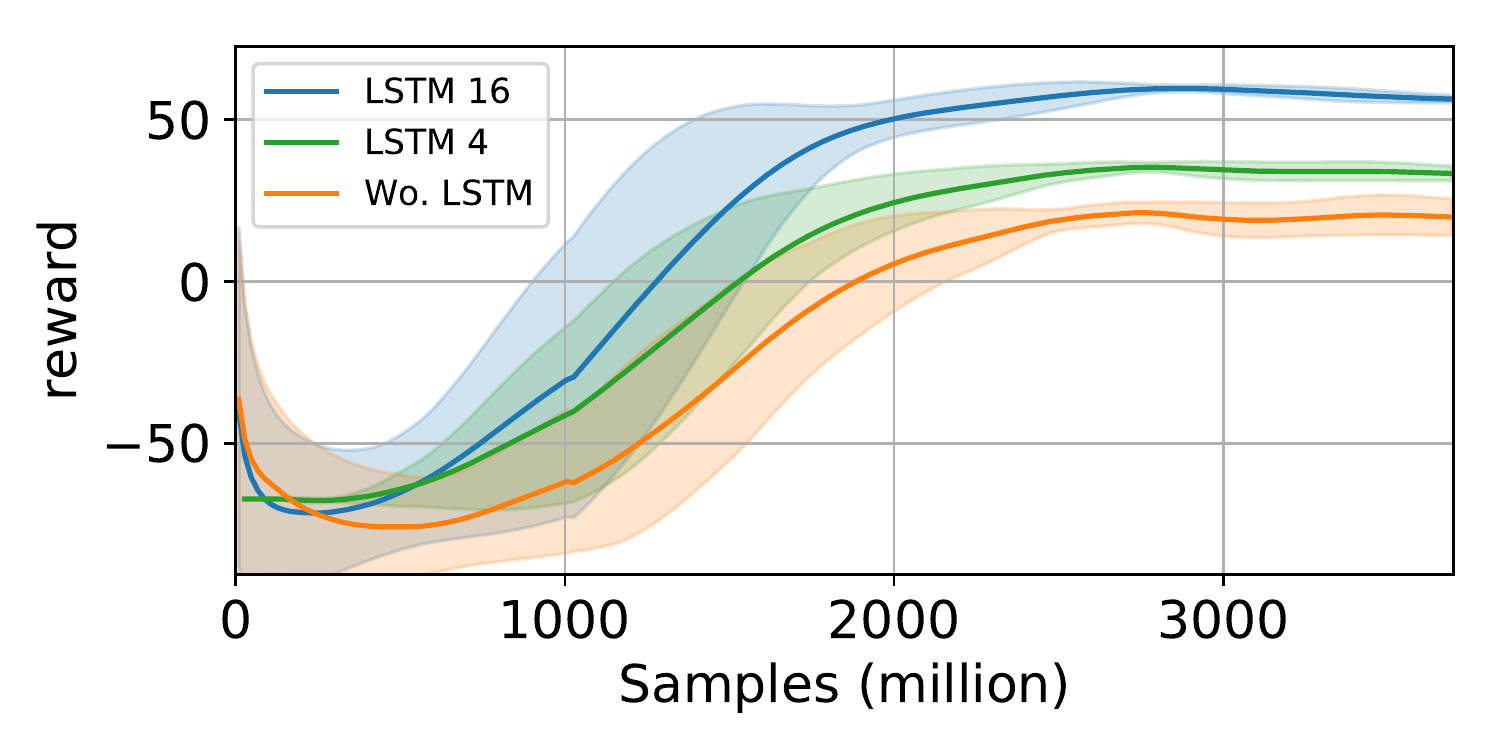} &
  \includegraphics[width=.31\linewidth]{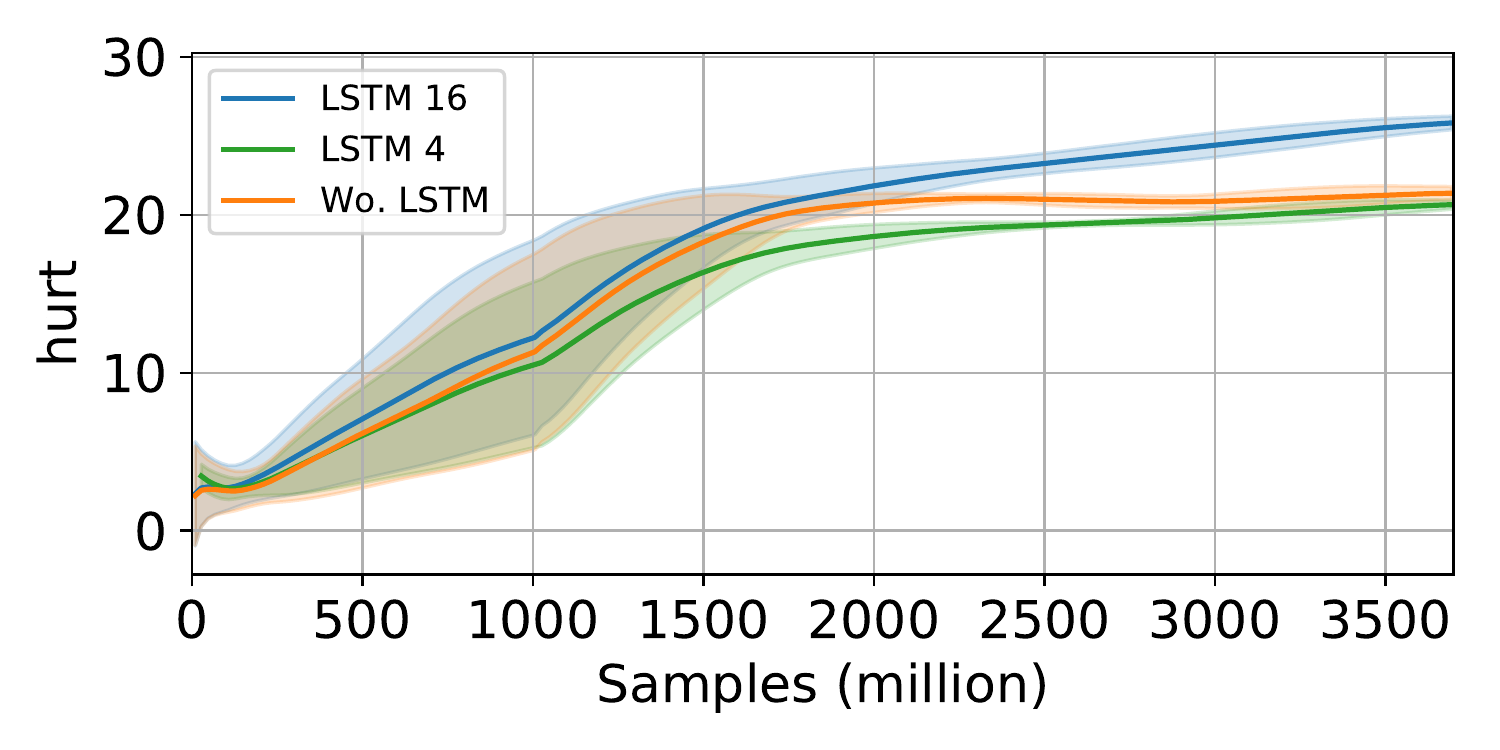} &
  \includegraphics[width=.31\linewidth]{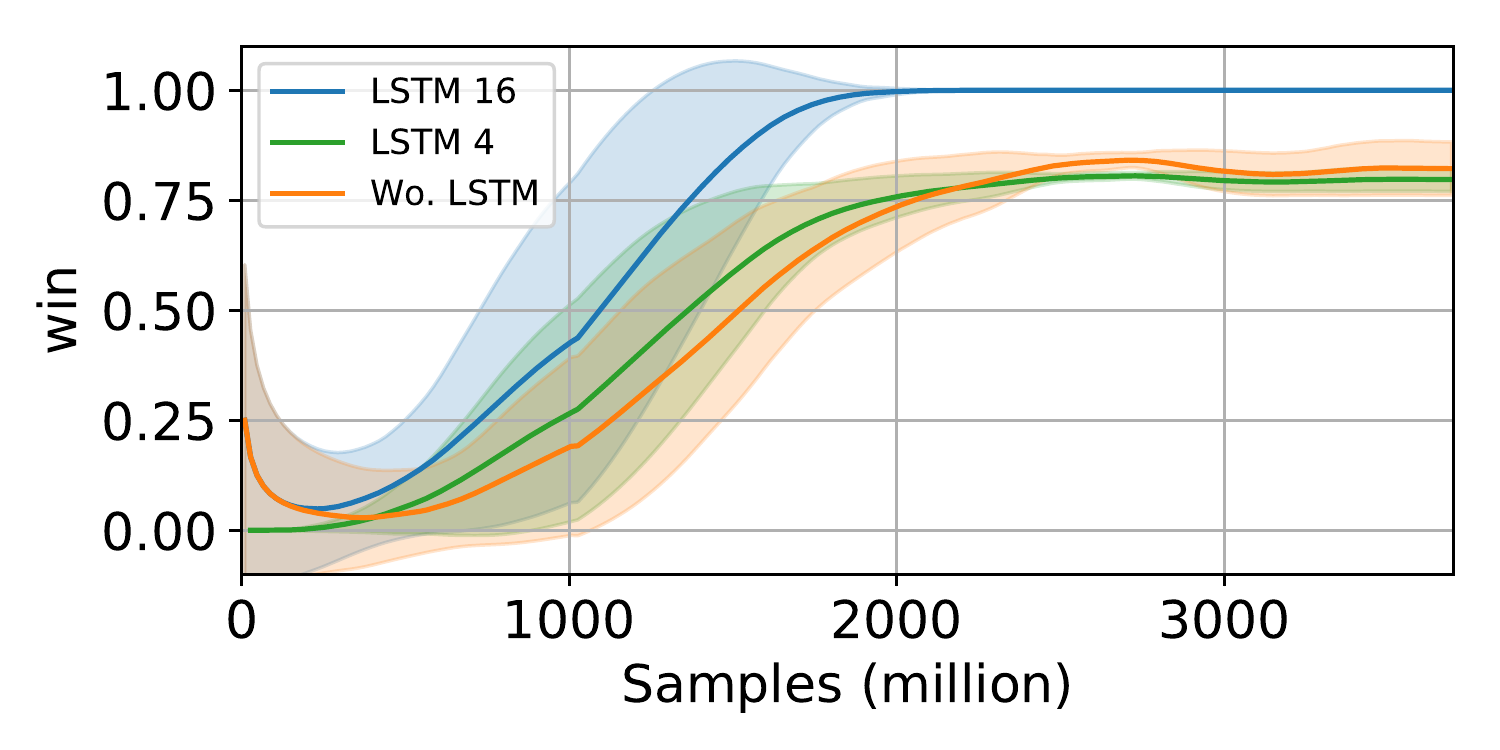}\\
  (a) Reward & (b) Hurt per frame  & c) Win rate \\
  \end{tabular}
 \caption{Different evaluation metrics on the \hokenv 1v1 mode for the ablation study on LSTM w.r.t. the number of training samples. Error bars represent standard deviation. With LSTMs, the model performs better against models without LSTMs; as the step size grows, the model performs better.}
    \label{fig:lstm}
\end{figure*}

To tackle the challenge from the partially observable environment, researchers often utilize LSTMs in the model network. In \hokenv, the action sequences of a hero, also called skill combos, are critical to creating severe and instant damage and should be incorporated with temporal information. We compared the models w/wo LSTMs, and the results are shown in Figure~\ref{fig:lstm}. As expected, the performance of models with LSTMs performs better than the model without LSTMs. 

\subsection{Evaluation under Elo} When a policy is trained against a fixed opponent, it may exploit its particular weaknesses and may not generalize well to other adversaries. We conducted an experiment to showcase this in which a first model A was trained against \bt. Then, another agent B was trained against a frozen version of agent A in the same scenario. While B managed to beat A consistently, its performance against built-in AI was not as good as Model A. The numerical results showing this lack of transitivity across the agents are presented in Table~\ref{tab:elo}.

\begin{table}[t!]
\caption{Performances of different agents under the evaluation of winning rate (with standard deviation) and Elo score. Although B could beat A among all the matches, B's winning rate is not as good as A's when competing with\bt. Instead, the Elo scores can reflect the relative capability.}
\centering
\label{tab:elo}
\begin{tabular}{cc}
\toprule
\textbf{Models in matches}            & \textbf{Winning rate} \\  \midrule
Model A vs. \bt & 0.91$\pm$0.03  \\
Model B vs. Model A          & 1.00$\pm$0.00  \\
Model B vs. \bt & 0.81$\pm$0.25  \\ \midrule
 \textbf{Models}           & \textbf{Elo Score} \\ \midrule
 Model A          &   2372  \\  
 Model B          &   1186   \\  
\bt &    0  \\   \bottomrule
\end{tabular}
\end{table}


To evaluate the ability of different agents, we suggest the necessity of using Elo score. We evaluate the Elo score for all the models, and the K factor in Elo score calculation is set to 200. As is shown in Table~\ref{tab:elo}, although Model B can perfectly beat Model A, its Elo score is less than A's. This is because because B is trained against a \emph{frozen} version of A and might overfit to A's weaknesses. The deviation of the winning rate (0.25) in B \emph{vs.} \bt is much larger than A \emph{vs.} \bt (0.03), which also indicates that B is worse than it shows in the mean of winning rates (0.91). This shows that Elo score considers the overall performance of one model against all the models, which is a more comprehensive metric than measuring the winning rate of one model against another model.

\subsection{Tasks on different heroes}
\begin{figure}[h!]
\centering
    \includegraphics[width=0.85\linewidth]{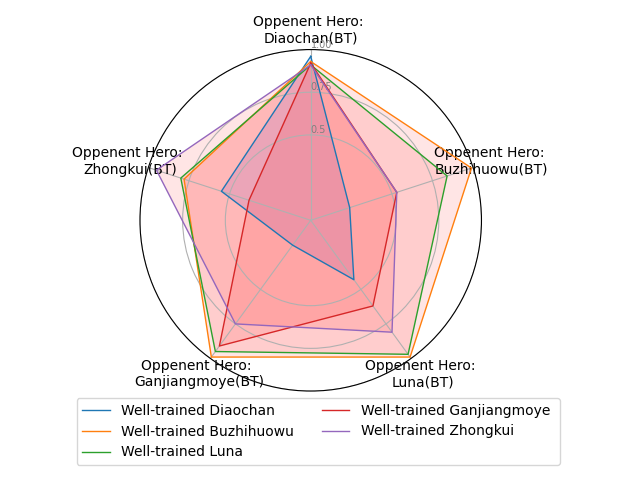}
  \caption{Generalization across different opponents with models trained from different heroes. Each line represents a well-trained model trained to control a target hero to compete with the same hero controlled by \bt. Each axis in the plot represents the testing task of different opponent hero to compete with for the same well-trained model. The value on the axis represents the winning rate under the task. A model trained on Luna and Buzhihuowu can generalize better to compete with four other heroes controlled by \bt.}
  \label{fig:more-heroes-oppo}
\end{figure}

\begin{figure}[h!]
\centering
    \includegraphics[width=0.85\linewidth]{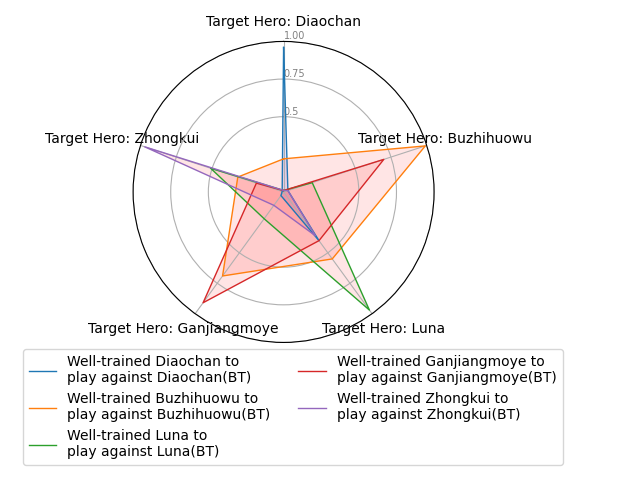}
  \caption{Generalization across different target heroes with models trained from different heroes. Each line represents a well-trained model trained to control a target hero to compete with the same hero controlled by \bt. Each axis in the plot represents the testing task of different target hero to control for the same well-trained model. The value on the axis represents the winning rate under the task. A model trained on Buzhihuowu can generalize better than models trained on other heroes to control different heroes.}
  \label{fig:more-heroes-target}
\end{figure}

\begin{figure}[h!]
\centering
    \includegraphics[width=0.99\linewidth]{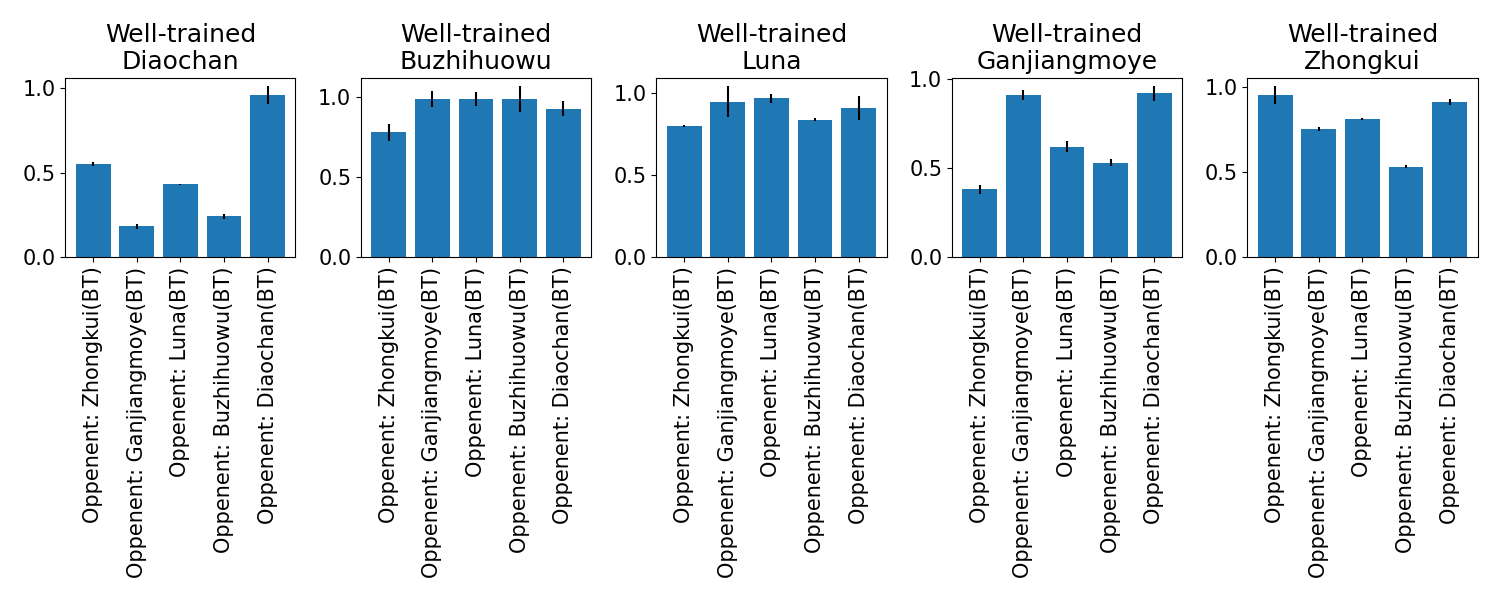}
  \caption{Bar charts of Figure~\ref{fig:more-heroes-oppo}.}
  \label{fig:more-heroes-oppo-bar}
\end{figure}

\begin{figure}[h!]
\centering
    \includegraphics[width=0.99\linewidth]{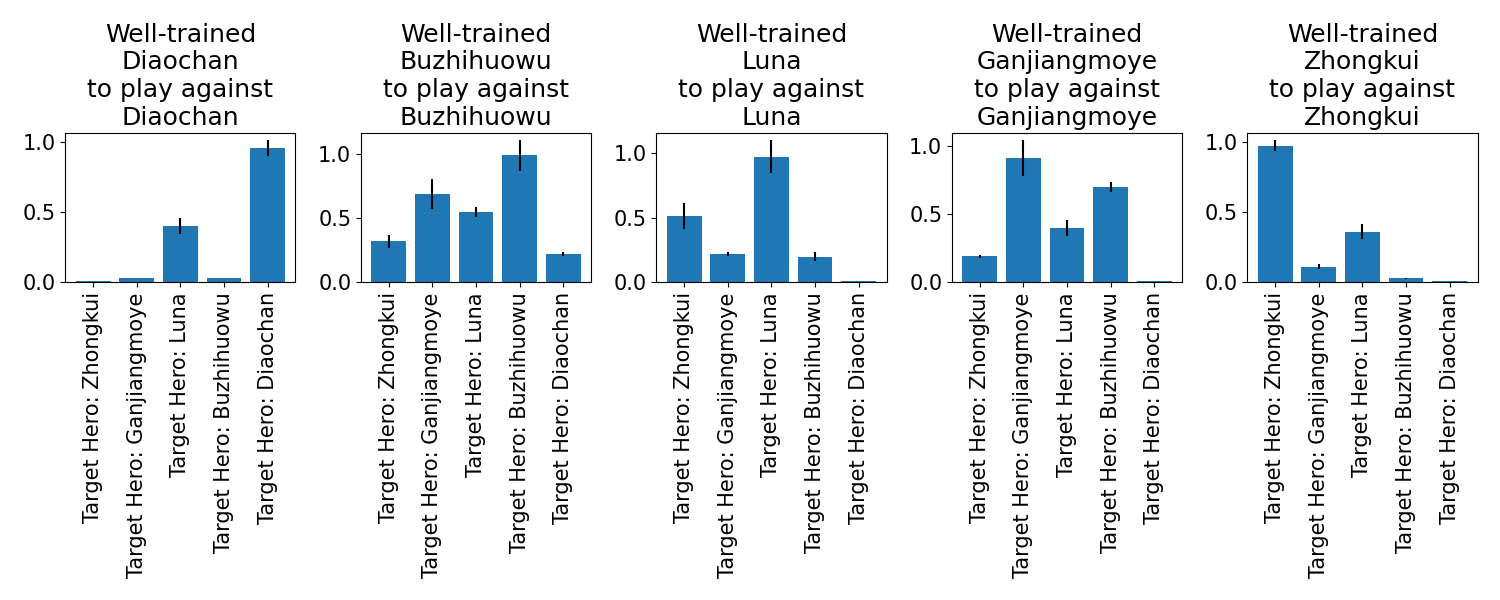}
  \caption{Bar charts of Figure~\ref{fig:more-heroes-target}.}
  \label{fig:more-heroes-target-bar}
\end{figure}

In Section~\ref{sec:generalization}, we chose the task "Diaochan vs. Diaochan" as the primary training task, while other feasible training tasks are reasonable options. Here we provide 5 different heroes as the primary training tasks and test their generalization ability across both opponents and targets with direct transfer. The results are shown in Figure~\ref{fig:more-heroes-oppo} and Figure~\ref{fig:more-heroes-target}. \rebuttal{The corresponding bar charts are provided as well in Figure~\ref{fig:more-heroes-oppo-bar} and Figure~\ref{fig:more-heroes-target-bar}.}

\end{document}